%% file: SPM_R2.tex
\newif\ifsingle
\singletrue 

\newif\ifFullVersion

\ifsingle
\documentclass[11pt,draftcls,onecolumn,journal]{IEEEtran}

\else		
\documentclass[10pt,final, twocolumn]{IEEEtran}
\fi

\input{Preamble}
\definecolor{NewColor}{rgb}{0,0,0}

\IEEEoverridecommandlockouts 

\ifsingle
\newcommand{\figWidth}{0.6\columnwidth}
\newcommand{\boxWidth}{\linewidth}
\else
\newcommand{\figWidth}{\linewidth}
\newcommand{\boxWidth}{2\linewidth}
\fi

\usepackage[all=normal,paragraphs=tight,floats=normal,mathspacing=normal,wordspacing=tight,charwidths=tight,mathdisplays=normal,leading=normal]{savetrees}

\title{AI-Aided Kalman Filters}
\author{
{
 Nir Shlezinger,~\IEEEmembership{Senior Member,~IEEE},
    Guy Revach,~\IEEEmembership{{Senior Member},~IEEE},
    Anubhab Ghosh,~\IEEEmembership{{Student Member},~IEEE},
    Saikat Chatterjee,~\IEEEmembership{Senior Member,~IEEE},
    Shuo Tang,~\IEEEmembership{Student Member,~IEEE},
    Tales Imbiriba,~\IEEEmembership{Member,~IEEE},
    Jindrich Dunik,~\IEEEmembership{Senior Member,~IEEE},
    Ondrej Straka,~\IEEEmembership{Member,~IEEE},
    Pau Closas,~\IEEEmembership{Senior Member,~IEEE},
    and Yonina C. Eldar,~\IEEEmembership{Fellow,~IEEE}
}
\thanks{
		 N. Shlezinger is with the School of ECE, Ben-Gurion University of the Negev, Beer-Sheva, Israel (e-mail: nirshl@bgu.ac.il).
		G. Revach is with the Institute for Signal and Information Processing, D-ITET, ETH Zürich, Switzerland (e-mail: grevach@ethz.ch).
		 A. Ghosh and S. Chatterjee are with the School of EE and CS, KTH Royal Institute of Technology, Stockholm, Sweden (e-mail: \{anubhabg; sach\}@kth.se).
        J. Dunik and O. Straka are with the Faculty of Applied Science, University of West Bohemia, Czech Republic (email: \{dunikj; straka30\}@kky.zcu.cz). T. Imbiriba is with the Dept. of CS, University of Massachusetts Boston, Boston, MA, USA (email: tales.imbiriba@umb.edu).
       S. Tang and P. Closas are with the Dept. of ECE, Northeastern University, Boston, MA, USA (email: \{tang.shu; closas\}@northeastern.edu). 
		 Y. C. Eldar is with the Math and CS Faculty, Weizmann Institute of Science, Rehovot, Israel (e-mail:  yonina.eldar@weizmann.ac.il). 
This work has been partially supported by  the European Research Council (ERC) under the ERC starting grant nr. 101163973 (FLAIR), by the National Science Foundation under Awards ECCS-1845833 and CCF-2326559, and by the Ministry of Education, Youth and Sports of the Czech Republic under project ROBOPROX - Robotics and Advanced Industrial Production
CZ.02.01.01/00/22\_008/0004590.}
 \vspace{-0.5cm}
} 

\begin{document}

\maketitle
\pagestyle{plain}
\thispagestyle{plain}

	\vspace{-0.75cm}

\begin{abstract}
\label{sec:abstract}
The \ac{kf} and its variants are among the most celebrated algorithms in signal processing. These methods are used for state estimation of dynamic systems by relying on mathematical representations in the form of simple \ac{ss} models, which may be crude and inaccurate descriptions of the underlying dynamics. Emerging data-centric \ac{ai} techniques tackle these tasks using \acp{dnn}, which are model-agnostic. Recent developments illustrate the possibility of fusing \acp{dnn} with classic Kalman-type filtering, obtaining systems that learn to track in partially known dynamics.  This article provides a tutorial-style overview of design approaches for incorporating \ac{ai} in aiding \ac{kf}-type algorithms. We review both generic and dedicated \ac{dnn} architectures suitable for state estimation and provide a systematic presentation of techniques for fusing \ac{ai} tools with \acp{kf} and for leveraging partial \ac{ss} modeling and data, categorizing design approaches into {\em task-oriented} and {\em \ac{ss} model-oriented}. The usefulness of each  approach in preserving the individual strengths of model-based \acp{kf} and data-driven \acp{dnn} is investigated in a qualitative and quantitative study  (whose  code is publicly available), illustrating the gains of hybrid model-based/data-driven designs.  We also discuss existing challenges and future research directions that arise from fusing \ac{ai} and Kalman-type algorithms. 
\end{abstract} 
\acresetall

\section{Introduction}
\label{sec:introduction} 
 The Apollo program, which successfully landed the first humans on the moon, is still considered one of mankind’s greatest achievements. Among the technological innovations that played a role in the success of the Apollo program is a filtering method based on the extended version of an algorithm developed by Rudolf Kalman in the late 1950s \cite{kalman1960new}, which was used to track and estimate the trajectory of the spaceship. This algorithm, known as the \ac{kf}, and its extension into the \ac{ekf}~\cite{schmidt1981kalman}, provide an accurate and reliable real-time trajectory estimation while still being simple to implement and applicable on the hardware-limited navigation computer of the Apollo spaceship~\cite{grewal2010applications}. To date, the \ac{kf} is among the most celebrated algorithms in signal processing and electrical engineering at large, with numerous applications including radar
, biomedical systems
, vehicular technology
, navigation, 
 and wireless communications.

The \ac{kf} is a {\em model-based} method, namely, it leverages mathematical parametric representations of the environment. Specifically, the \ac{kf} and its variants \cite{durbin2012time} rely on the ability to describe the underlying dynamics as a \ac{ss} model. Such model-based designs have several core advantages when the dynamics are faithfully captured using a known and tractable \ac{ss} representation: 
$(i)$ they can achieve {\em optimal performance}. For example, the \ac{kf} achieves the \textcolor{NewColor}{minimum} \ac{mse}  for linear Gaussian \ac{ss} models; 
$(ii)$ model-based methods operate with {\em controllable and often reduced complexity}. In fact, the \ac{kf} and its variants are implemented on devices such as sensors and mobile systems, where they operate in real-time using limited computational and power resources;
$(iii)$ decisions made by the \ac{kf} and its variants are {\em interpretable}, in the sense that their internal features have concrete  meaning and their reasoning can be explained based on the  observations and the \ac{ss} model;
$(iv)$ they are inherently {\em adaptive}, as changes in the \ac{ss} model are naturally incorporated into the operation; 
and $(v)$ they reliably characterize {\em uncertainty} in their estimate, providing the error covariance alongside their estimates~\cite{sarkka2023bayesian}.

A key characteristic of \ac{kf}-type algorithms is their reliance on knowledge of the underlying dynamics and specifically, on the ability to accurately capture these dynamics using a known and tractable \ac{ss} representation. A common approach is to rely on simplifying assumptions (e.g., linear systems, Gaussian mutually and temporally independent process and measurement noises, etc.) that make models understandable and the associated algorithms computationally efficient, and then use data to estimate the unknown parameters. However, simple models frequently fail to represent some of the nuances and subtleties of dynamic systems and their associated signals. Practical applications are thus often required to operate with partially known \ac{ss} models. Furthermore, the performance and reliability of \ac{kf}-type algorithms degrade considerably when using a postulated \ac{ss} model that deviates from the true nature of the underlying system. 

The unprecedented success of \ac{ml}, and particularly deep learning, as the enabler technology for \ac{ai} in areas such as computer vision has initiated a general mindset geared towards data. It is now quite common practice to replace simple principled models with data-driven pipelines, employing \ac{ml} architectures based on \acp{dnn} that are trained with massive volumes of labeled data. \Acp{dnn} can be trained end-to-end without relying on analytical approximations, and therefore, they can operate in scenarios where analytical models are unknown or highly complex \cite{goodfellow2016deep}. With the growing popularity of deep learning, recent years have witnessed the design of various \acp{dnn} for tasks associated with \ac{kf}-type algorithms, including, e.g., the application of \acp{dnn} for state estimation \cite{becker2019recurrent, krishnan2017structured} and control  \cite{kuutti2020survey}. 

\ac{ai} systems can learn from data to track dynamic systems without relying on full knowledge of underlying \ac{ss} models. However, replacing \ac{kf}-type algorithms with deep architectures gives rise to several core challenges: 
\textcolor{NewColor}{$(i)$ the {\em computational burden} of training and utilizing highly parametrized \acp{dnn}, as well as the fact that {\em massive data sets} are typically required for their training, often constitute major drawbacks in various signal-processing applications. This limitation is particularly relevant for hardware-constrained devices (such as mobile systems and sensors) whose ability to utilize highly parametrized \acp{dnn} is typically limited~\cite{chen2019deep}; 
$(ii)$  due to the complex and generic structure of \acp{dnn}, it is often {\em challenging to understand}  how they obtain their predictions, or track the rationale leading to their decisions; and $(iii)$ \acp{dnn}  typically struggle to adapt to variations in the data distribution and are limited in their capability to provide an estimation of uncertainty.} 

The challenges outlined above gave rise to a growing interest in using \ac{ai} not to replace \ac{kf}-type algorithms, but to empower them, as a form of hybrid model-based/data-driven design~\cite{shlezinger2020model}. 
Various approaches have been proposed for combining \acp{kf} and deep learning including training \acp{dnn} to map data into features that obey a known \ac{ss} model and can be tracked with a \ac{kf}
~\cite{klushyn2021latent, coskun2017long}; employing deep models for identifying \ac{ss} models to be used for \ac{kf}-based tracking
~\cite{gedon2021deep, imbiriba2023augmented}
and converting the \ac{kf} into an \ac{ml} model that can be trained in a supervised
~\cite{revach2022kalmannet,buchnik2023latent,welling2021hkf,satorras2019combining} 
or unsupervised~\cite{ghosh2023danse,revach2021unsupervised} manner. Such \ac{ai}-empowered \acp{kf} were already in various signal processing applications, 
ranging from brain-machine interface, acoustic echo cancellation, financial monitoring, beam tracking in wireless systems, and drone-based monitoring systems ~\cite{cubillos2025exploringSHORT, aspragkathos2023event, milstein2024neural}.  
These recent advances in combining \ac{ai} and \acp{kf}, along with their implications on emerging technologies, motivate a systematic presentation of the different design approaches, as well as their associated signal processing challenges. 

In this article, we provide a tutorial-style overview of the design approaches and potential benefits of combining deep learning tools with \ac{kf}-type algorithms. We refer to this as `AI-aided Kalman Filters'. While highlighting the potential benefits of \ac{ai}-aided \acp{kf}, we also mention the main signal processing challenges that arise from such hybrid designs. 
For this goal, we begin by reviewing the fundamentals of KF that are relevant to understanding its fusion with \ac{ai}. We briefly describe its core statistical representation -- the \ac{ss} model -- and formulate the mathematical steps for filtering and smoothing. We then discuss the pros and cons associated with \ac{kf}-type algorithms and pinpoint the main challenging aspects that motivate the usage of \ac{ai} tools. 
Next we shift our focus to deep learning techniques that are suitable for processing time sequences, briefly reviewing both generic time-sequence architectures, such as \acp{rnn} and attention mechanisms, and proceeding to  \ac{dnn} architectures that are inspired by \ac{ss} representations~\cite{gu2021combining} and by \ac{kf} processing flow~\cite{becker2019recurrent}. We discuss the gains offered by such data-driven techniques, while also highlighting their limitations, in turn indicating the potential of combining \ac{ai} techniques with classic \ac{ss} model-based \ac{kf}-type methods.

The bulk of the article is dedicated to presenting design approaches that combine deep learning techniques with \ac{kf}-type processing based on a partial mathematical representation of the dynamics. We categorize existing design approaches, drawing inspiration from the \ac{ml} paradigms of {\em discriminative} (task-oriented) learning and {\em generative} (statistical model-oriented) learning~\cite{shlezinger2022discriminative}. Accordingly, the first part is dedicated to \ac{ai}-augmented \acp{kf} via task-oriented learning, presenting in detail candidate approaches for converting  \ac{kf}-type filtering into an \ac{ml} architecture via \ac{dnn}-augmentation. These include designs that employ an external \ac{dnn}, either sequentially~\cite{zhou2020kfnet} or in parallel~\cite{satorras2019combining}, as well as \acp{dnn} augmented into a \ac{kf}-type algorithm, e.g., for \ac{kg} computation~\cite{revach2022kalmannet}. 
We then proceed to detail \ac{ss} model-oriented \ac{ai}-aided \acp{kf} designs. These can be viewed as a form of \ac{ai}-aided system identification, i.e., techniques that utilize \acp{dnn} in the process of identifying \ac{ss} models, which can then be used by model-based \ac{kf}-type tracking~\cite{gedon2021deep, imbiriba2023augmented,jouaber2021nnakf}. We discuss several approaches with various structures connecting physics-based and data-based model components, such as incremental, hybrid, or fixed \ac{ss} structures.

To highlight the gains of each approach for designing \ac{ai}-aided \acp{kf}, and capture the interplay between the different algorithms and conventional model-based or contemporary data-driven methods, we provide a comparative study. We begin with a qualitative comparison, pinpointing the conceptual differences between the design approaches in terms such as the level of domain knowledge required, the type of data needed, and flexibility. We also provide a quantitative comparison, where we evaluate representative methods from the presented design approaches in a setting involving the tracking of the challenging Lorenz attractor.  
The article concludes with a discussion of open challenges and future research directions. 


{\bf Notations:} We use boldface lowercase for vectors, e.g., $\myVec{x}$ and boldface uppercase letters, e.g., $\myMat{X}$ for matrices. 
For a time sequence $\myVec{x}_t$ and time indices $t_1 \leq t_2$, we use the abbreviated form $\myVec{x}_{t_1:t_2}$ for the set of  variables $\{\myVec{x}_t\}_{t=t_1}^{t_2}$. 
We use $\mathcal{N}(\myVec{\mu},\myMat{\Sigma})$ for the multivariate Gaussian distribution with mean $\myVec{\mu}$ and covariance $\myMat{\Sigma}$, while $\pdf(\cdot)$ is a probability density function, while $\Vert \myVec{x} \Vert^2_{\myMat{C}}$ is the squared $\ell_2$ norm of $\myVec{x}$ weighted by the matrix $\myMat{C}$, i.e. $\Vert \myVec{x} \Vert^2_{\myMat{C}} = \myVec{x}^{\top} \myMat{C} \myVec{x}$. The operations $(\cdot)^\top$ and  $\| \cdot \|_2$  are the transpose and  $\ell_2$ norm,  respectively. 
\section{Fundamentals of Kalman Filtering} \label{sec:fund_KF}

In this section, we review some basics of Kalman-type filtering. We commence with reviewing \ac{ss} models, after which we recall the formulation of the \ac{kf} and \ac{ekf}. We conclude this section by highlighting the challenges that motivate its augmentation with deep learning.

\subsection{State-Space Models}
\label{ssec:fund_ss}

\ac{ss} models are a class of mathematical models that describe the probabilistic behavior of dynamical systems. They constitute the fundamental framework for formulating a broad range of engineering problems in the areas of signal processing, control, and communications, and they are also widely used in environment studies, economics,  and many more. The core of \ac{ss} models lies in the representation of the dynamics of a system using a latent state variable that evolves over time while being related to the observations made by the system. 

{\bf Generic \ac{ss} Models}: Focusing on discrete-time formulations of continuous-valued variables, \ac{ss} models represent the interplay between the observations at time  $t$, denoted $\myVec{y}_t$, an input signal $\myVec{u}_t$, and a  state capturing the system dynamics $\myVec{x}_t$. In general, \ac{ss} models consist of: $(i)$ a {\em state evolution} model of the form 
\begin{subequations}
    \label{eq:ssModel}
\begin{equation}
    \label{eq:ssModelState}
    \myVec{x}_{t+1} = \tilde{f}_t(\myVec{x}_{t}, \myVec{u}_{t}, \myVec{v}_{t}),
\end{equation}
representing how the state evolves in time, \textcolor{NewColor}{with $\myVec{v}_t$ being the {\em process noise}}; and $(ii)$ an observation model 
\begin{equation}
    \label{eq:ssModelObs}
    \myVec{y}_{t} = \tilde{h}_t(\myVec{x}_{t}, \myVec{w}_{t}),
\end{equation}
\end{subequations}
which relates the observations and the current system state, \textcolor{NewColor}{where $\myVec{w}_t$ represents the {\em measurement noise}}. While the mappings $\tilde{f}_t(\cdot)$ and $\tilde{h}_t(\cdot)$ are deterministic, stochasticity is induced by the temporally independent noises $\myVec{v}_t$ and $\myVec{w}_t$, and by the distribution of the initial state $\myVec{x}_0$. 

{\bf Gaussian \ac{ss} Models}: A common special case of the above generic model is that \textcolor{NewColor}{referred to as} the {\em linear Gaussian \ac{ss} model}, used by the celebrated \ac{kf}. In this \textcolor{NewColor}{special case of \eqref{eq:ssModel}}, the state evolution takes the form 
\begin{subequations}
    \label{eq:LinssModel}
\begin{equation}
    \label{eq:LinssModelState}
\myVec{x}_{t+1} = \myMat{F}_t\myVec{x}_{t} + \myMat{G}_t \myVec{u}_{t} + \myVec{v}_{t},
\end{equation}
and the observation model is given by 
\begin{equation}
\label{eq:LinssModelObs}
\myVec{y}_t=\myMat{H}_t\myVec{x}_t+\myVec{w}_t,
\end{equation}
\end{subequations}
where $\myMat{F}_t,\myMat{G}_t,\myMat{H}_t$ are matrices of appropriate dimensions, and $\myVec{v}_t$ and $\myVec{w}_t$ are temporally and mutually independent  zero-mean Gaussian signals,  with covariances $\myMat{Q}_t$ and $\myMat{R}_t$, respectively. Namely, $\myVec{v}_t\sim \mathcal{N}(\myVec{0},\myMat{Q}_t)$ and $\myVec{w}_t\sim \mathcal{N}(\myVec{0},\myMat{R}_t)$, while $\myVec{v}_t$ is independent of $\myVec{w}_t$, and both are independent of $\myVec{v}_\tau$ and $\myVec{w}_\tau$ for any $\tau \neq t$. The initial state  is independent of the noises and  assumed to obey $\myVec{x}_0 \sim \mathcal{N}(\hat{\myVec{x}}_0,\hat{\myMat{\Sigma}}_0)$, with known $\hat{\myVec{x}}_0$, $\hat{\myMat{\Sigma}}_0$.

Similar models to \eqref{eq:LinssModel} are frequently utilized for settings characterized by non-linear transformations. In  non-linear \ac{ss} models with additive Gaussian noises (termed henceforth as non-linear Gaussian), \eqref{eq:ssModel} takes the form 
\begin{subequations}
    \label{eq:NLssModel}
    \begin{align}
    \myVec{x}_{t+1} &= f_t(\myVec{x}_{t}, \myVec{u}_{t}) + \myVec{v}_{t}, \label{eq:NLssModelState}\\
    \myVec{y}_{t} &= h_t(\myVec{x}_{t}) + \myVec{w}_{t}, \label{eq:NLssModelObs}
\end{align}  
\end{subequations}
where the noise signals   are as in \eqref{eq:LinssModel}, i.e.,  temporally independent with $\myVec{v}_t\sim \mathcal{N}(\myVec{0},\myMat{Q}_t)$ and $\myVec{w}_t\sim \mathcal{N}(\myVec{0},\myMat{R}_t)$.  

	{\bf Tasks}: \ac{ss} models are mostly associated with two main families of tasks. The first is {\em state estimation}, which deals with the recovery of the state variable $\myVec{x}_t$ based on a set of observations $\{\myVec{y}_\tau\}$, i.e., an open-loop system where the input $\myVec{u}_t$ is either absent or not controlled. Common state estimation tasks include~\cite{durbin2012time}
     $(i)$ {\em Filtering:} estimate $\myVec{x}_t$ from $\myVec{y}_{1:t}$; and
     $(ii)$ {\em Smoothing:} estimate  $\myVec{x}_{1:T}$ from $\myVec{y}_{1:T}$ for some $T>0$. 
 Additional related tasks are prediction, input recovery, and imputation. State estimation plays a key role in applications that involve tracking, localization, and denoising. 
 
 The second family of \ac{ss} model-based tasks are those that deal with {\em stochastic control} (closed-loop) policies. In this family, the \ac{ss} framework is used to select how to set the input variable $\myVec{u}_t$ based on, e.g., past measurements $\myVec{y}_{1:t}$. Such tasks are fundamental in robotics, vehicular systems, and aerospace engineering. As stochastic control policies often employ state estimation schemes followed by state regulators, 
 we focus in this article on state estimation (i.e., the first family of tasks). 

\subsection{Kalman Filtering}
\label{ssec:fund_KF_EKF}
The representation of dynamic systems via \ac{ss} models gives rise to some of the most celebrated algorithms in signal processing, particularly the family of Kalman-type filters. To describe these, we henceforth focus on {\em state estimation}. Therefore, for convenience, we omit the input signal $\myVec{u}_t$ from the following relations (as it can also be absorbed into the process noise as a known bias term).

{\bf \ac{kf}}: 
The \ac{kf} is the \textcolor{NewColor}{minimum} \ac{mse} estimator for the filtering task in linear Gaussian \ac{ss} models, i.e., estimating $\myVec{x}_t$ from $\myVec{y}_{1:t}$ when these are related via \eqref{eq:LinssModel}. In  time step~$t$, the \ac{kf} estimates  $\myVec{x}_t$ using  the previous estimate $\hat{\myVec{x}}_{t-1}$ as a sufficient statistic and the observed $\myVec{y}_{t}$, thus its complexity does not grow in time.

The \ac{kf} updates its estimates of the first- and second-order statistical moments of the state, which at time $t$ are denoted by $\hat{\myVec{x}}_{t\given{t}}$ and $\myMat{\Sigma}_{t\given{t}}$, respectively. For $t=0$, these moments are initialized to those of the initial state, namely $\hat{\myVec{x}}_0$ and $\hat{\myMat{\Sigma}}_0$, while for $t>0$ the estimates are obtained via a two-step procedure:
\begin{enumerate}
    \item {\em Prediction:} The first step \emph{predicts} the first- and second-order statistical moments of current \textit{a priori} state and observation,  based on the previous \textit{a posteriori} estimate. Specifically, at time $t$ the predicted moments of the state are computed via
    \begin{subequations}\label{eqn:predict_evol}
    \begin{align}\label{eqn:evol_1}
    \hat{\myVec{x}}_{t\given{t-1}} &= 
    \myMat{F}_{t-1}\cdot{\hat{\myVec{x}}_{t-1\given{t-1}}},\\\label{eqn:evol_2}
    \myMat{\Sigma}_{t\given{t-1}} &=
    {\myMat{F}_{t-1}}\cdot\myMat{\Sigma}_{t-1\given{t-1}}\cdot\myMat{F}_{t-1}^\top+\myMat{Q}_{t-1}.
    \end{align}
    \end{subequations}
    The predicted moments of the observations  are computed as
    \begin{subequations}\label{eqn:predict_obs}
    \begin{align}\label{eqn:obs_1}
    \hat{\myVec{y}}_{t\given{t-1}} &=
    \myVec{H}_t\cdot\hat{\myVec{x}}_{t\given{t-1}},\\\label{eqn:obs_2}
    {\myMat{S}}_{t\given{t-1}} &=
    {\myMat{H}_t}\cdot\myMat{\Sigma}_{t\given{t-1}}\cdot\myMat{H}_t^\top+\myMat{R}_t.
    \end{align}
    \end{subequations}
    
    \item {\em Update:} The predicted \textit{a priori} moments are updated using the current observation $\myVec{y}_t$ into the \textit{a posteriori} state moments. The updated moments are computed as
    \begin{subequations}\label{eqn:update}
    \begin{align}\label{eqn:update1}
    \hat{\myVec{x}}_{t\given{t}}&=
    \hat{\myVec{x}}_{t\given{t-1}}+\Kgain_{t}\cdot\Delta\myVec{y}_t,\\\label{eqn:update2}
    {\myMat{\Sigma}}_{t\given{t}}&=
    {\myMat{\Sigma}}_{t\given{t-1}}-\Kgain_{t}\cdot{\myMat{S}}_{t\given{t-1}}\cdot\Kgain^{\top}_{t}.
    \end{align}
    \end{subequations}
    Here, $\Kgain_{t}$ is the \ac{kg}, and it is given by
    \begin{equation}\label{eq:FWGain}
    \Kgain_{t}={\myMat{\Sigma}}_{t\given{t-1}}\cdot{\myMat{H}_t^\top}\cdot{\myMat{S}}^{-1}_{t\given{t-1}}.
    \end{equation}
    The term $\Delta\myVec{y}_t\triangleq \myVec{y}_t-\hat{\myVec{y}}_{t\given{t-1}}$ is then defined as the {\em innovation}, representing the difference between the predicted observation and the observed value.
\end{enumerate}
 The output of the \ac{kf} is the estimated state, i.e., $\hat{\myVec{x}}_t \equiv \hat{\myVec{x}}_{t\given{t}}$, and the error covariance is ${\myMat{\Sigma}}_{t} =   {\myMat{\Sigma}}_{t\given{t}}$. 
Various proofs are provided in the literature for the \ac{mse} optimality of the \ac{kf}~\cite{durbin2012time}. A relatively compact way to prove this follows from the more general Bayes' rule and Chapman-Kolmogorov equation, recalled in the box entitled {\em From Chapman-Kolmogorov to the \ac{kf}} on Page~\pageref{Box:ChapmanKolmo}. 

\begin{tcolorbox}[float*=t,
    width=\boxWidth,
	toprule = 0mm,
	bottomrule = 0mm,
	leftrule = 0mm,
	rightrule = 0mm,
	arc = 0mm,
	colframe = myblue,
	colback = mypurple,
	fonttitle = \sffamily\bfseries\large,
	title = From Chapman-Kolmogorov to the \ac{kf}]	
	\label{Box:ChapmanKolmo}
Consider a generic \ac{ss} model as in \eqref{eq:ssModel} without an input signal $\myVec{u}_t$. By Bayes' rule and the Markovian nature of  \eqref{eq:ssModel}, it holds that the conditional distribution of $\myVec{x}_t|\myVec{y}_{1:t}$ satisfies~\cite{barber2010graphical} 
\begin{subequations}
    \label{eqn:CK1}
	\begin{align}
	\pdf\left(\myVec{x}_t \given{\myVec{y}_{1:t}}\right) =\frac{\pdf\left(\myVec{y}_t|\myVec{x}_t\right)\pdf\left(\myVec{x}_t\given{\myVec{y}_{1:t-1}}\right)}{\pdf\left(\myVec{y}_t\given{\myVec{y}_{1:t-1}}\right)},
	\label{eqn:CK1a}
	\end{align}
 where 
	\begin{align}
	\pdf\left(\myVec{y}_t\given{\myVec{y}_{1:t-1}}\right)
&= \int \pdf(\myVec{y}_t| \myVec{x}_t) \pdf\left(\myVec{x}_t\given{\myVec{y}_{1:t-1}}\right) d\myVec{x}_t,
	\label{eqn:CK1b} \\
 \pdf\left(\myVec{x}_t\given{\myVec{y}_{1:t-1}}\right)
 &=  \int \pdf(\myVec{x}_t|\myVec{x}_{t-1}) \pdf\left(\myVec{x}_{t-1}\given{\myVec{y}_{1:t-1}}\right) d\myVec{x}_{t-1}.
	\label{eqn:CK1c}
	\end{align}
 \end{subequations} 
 Combining~\eqref{eqn:CK1a}-\eqref{eqn:CK1b} with~\eqref{eqn:CK1c}, referred to as the {\em Chapman-Kolmogorov} equation for \ac{ss} models~\cite[Ch. 4.2]{sarkka2023bayesian}, describes how  the posterior  $\pdf\left(\myVec{x}_{t-1}\given{\myVec{y}_{1:t-1}}\right)$ is recursively updated into $\pdf\left(\myVec{x}_{t}\given{\myVec{y}_{1:t}}\right)$. 

	For the special case of a linear Gaussian \ac{ss} model as in \eqref{eq:ssModel}, all the considered variables are jointly Gaussian. Specifically,  if $\myVec{x}_{t-1}|\myVec{y}_{1:t-1}\sim \mathcal{N}\left(\hat{\myVec{x}}_{t-1|t-1}, \myMat{\Sigma}_{t-1|t-1}\right)$, then  \eqref{eqn:CK1c} implies that $\myVec{x}_t\given{\myVec{y}_{1:t-1}} \sim\mathcal{N}\left(\hat{\myVec{x}}_{t|t-1}, \myMat{\Sigma}_{t|t-1} \right)$ (computed via \eqref{eqn:predict_evol}), which together with \eqref{eqn:CK1b} indicates that $\myVec{y}_t\given{\myVec{y}_{1:t-1}} \sim \mathcal{N}\left(\hat{\myVec{y}}_{t|t-1}, \myMat{S}_{t|t-1}  \right)$ (computed via \eqref{eqn:predict_obs}). Combining these with \eqref{eqn:CK1a} reveals that  $\myVec{x}_t|\myVec{y}_{1:t}\sim \mathcal{N}\left(\hat{\myVec{x}}_{t|t}, \myMat{\Sigma}_{t|t}\right)$ with moments computed via~\eqref{eqn:update}. Accordingly, $\hat{\myVec{x}}_t$ computed by the \ac{kf} is exactly the conditioned expectation of $\myVec{x}_t$ conditioned on $\myVec{y}_{1:t}$, i.e., it is  \ac{mse} optimal.
\end{tcolorbox}

\smallskip
{\bf Extension to Smoothing}: 
While the \ac{kf} is formulated for the filtering task, it also naturally extends (while preserving its optimality) for smoothing tasks. A leading approach to realize a smoother is the \ac{rts} algorithm~\cite{sarkka2023bayesian},
which employs two subsequent recursive passes to estimate the state, termed {\em forward}  and {\em backward} passes. The forward pass is the standard \ac{kf}, while the backward pass refines the estimates for each time $t$ using future observations at time instants $\tau \in \{t+1,\ldots,T\}$.

The backward pass is similar in its structure to the update step in the \ac{kf}. 
\textcolor{NewColor}{For the time step $t=T$, it uses $\hat{\myVec{x}}_{T\given{T}}$ and ${\myMat{\Sigma}}_{T\given{T}}$ computed by the forward recursion of the \ac{kf}. Then, it carries out a backward recursion, where}
for each $t\in\{T-1,\ldots,1\}$, the forward belief is corrected with future estimates via
\begin{subequations}\label{eq:BW_update}
\begin{align}\label{eq:BW_update1}
\hat{\myVec{x}}_{t\given{T}}&=
\hat{\myVec{x}}_{t\given{t}}+\overleftarrow{\Kgain}_t\cdot (\hat{\myVec{x}}_{t+1\given{T}} -  \hat{\myVec{x}}_{t+1\given{t}}),\\\label{eq:BW_update2}
{\myMat{\Sigma}}_{t\given{T}}&={\myMat{\Sigma}}_{t|t} - \overleftarrow{\Kgain}_t\cdot 
({\myMat{\Sigma}}_{t+1\given{t}}-{\myMat{\Sigma}}_{t+1\given{T}})
\cdot \overleftarrow{\Kgain}_t^{\top}.
\end{align}
\end{subequations}
%
%
Here, $\overleftarrow{\Kgain}_t$ is the {backward} \ac{kg}, computed based on second-order statistical moments from the forward pass as 
\begin{equation}\label{eq:bw_gain}
\overleftarrow{\Kgain}_t={\myMat{\Sigma}}_{t|t} \cdot \myMat{F}_t^{\top}\cdot{\myMat{\Sigma}}_{t+1\given{t}}^{-1}.
\end{equation}
 The output of the \ac{rts} is the estimated state for every time instant $t\in \{1,\ldots,T\}$, with $\hat{\myVec{x}}_t \equiv \hat{\myVec{x}}_{t\given{T}}$, as well as the error covariance ${\myMat{\Sigma}}_{t} \equiv   {\myMat{\Sigma}}_{t\given{T}}$.  

\smallskip
{\bf Extensions to Non-Linear Models}: 
The \ac{kf} is \ac{mse} optimal for linear and Gaussian \ac{ss} models. For non-linear settings,  approaches vary between non-linear Gaussian models as in \eqref{eq:NLssModel}, and non-Gaussian settings. 

Filtering algorithms for state estimation in non-linear Gaussian \ac{ss} models are typically designed to preserve the linear operation of the \ac{kf} update step with respect to measurement. 
\color{NewColor}
The key challenge in approximating the operation of the \ac{kf} lies in propagation of the first- and second-order moments. Arguably the most common non-linear variant of the \ac{kf}, known as the \ac{ekf}, is based on local linearizations.
In the \ac{ekf}, prediction of the first-order moments in \eqref{eqn:evol_1} and \eqref{eqn:obs_1} is replaced with
\ifsingle
\begin{equation}\label{eqn:predict_NL} 
\hat{\myVec{x}}_{t\given{t-1}} = 
{f}_t\left({\hat{\myVec{x}}_{t-1 \given t-1}}\right), \quad 
\hat{\myVec{y}}_{t\given{t-1}} =
{h}_t\left(\hat{\myVec{x}}_{t\given{t-1}}\right).
\end{equation}
\else
\begin{subequations}\label{eqn:predict_NL}
\begin{align}\label{eqn:evol_NL}
\hat{\myVec{x}}_{t\given{t-1}} &= 
{f}_t\left({\hat{\myVec{x}}_{t-1 \given t-1}}\right),\\\label{eqn:obs_NL}
\hat{\myVec{y}}_{t\given{t-1}} &=
{h}_t\left(\hat{\myVec{x}}_{t\given{t-1}}\right).
\end{align}
\end{subequations}
\fi 
 For the second-order moments, the matrices $\myMat{F}_t$ and $\myMat{H}_t$ in the \ac{kf} formulations are respectively replaced with the Jacobian matrices of $f_t(\cdot)$ and $h_t(\cdot)$, i.e.,  
\begin{subequations}\label{eqn:update_NL}
\begin{align}
\label{eqn:update_NL1}
\hat{\myMat{F}}_t &= \nabla_{\myVec{x}_{t-1}}f_{t-1} ({\myVec{x}}_{t-1})|_{{\myVec{x}}_{t-1}=\hat{\myVec{x}}_{t-1|t-1}}, \\
\hat{\myMat{H}}_t &= \nabla_{\myVec{x}_{t}}h_t ({\myVec{x}}_{t})|_{{\myVec{x}}_{t}=\hat{\myVec{x}}_{t|t-1}}.
%
\label{eqn:update_NL2}
\end{align}
\end{subequations}
Alternatively, the propagation of first- and second-order moments can be approximated using the unscented transform, resulting in the \ac{ukf},
or using the cubature
and Gauss-Hermite deterministic or stochastic quadrature
rules.
\color{black}

The \ac{ekf} and \ac{ukf} are approximations of the \ac{kf} designed for non-linear \ac{ss} models \textcolor{NewColor}{with Gaussian distributed noise}. When the \ac{ss} \textcolor{NewColor}{representation has non-Gaussian noise}, state estimation algorithms typically aim at recursively updating the posterior distribution via the Chapman-Kolmogorov relation, without resorting to its Gaussian special case utilized by the \ac{kf} (See box on page~\pageref{Box:ChapmanKolmo}). Leading algorithms that operate in this manner include the family of particle filters, that are based on sequential sampling;
Gaussian sum filters, based on Gaussian sum representation of all densities;
and point-mass filters, numerically solving the Bayesian relation in a typically rectangular grid.


\subsection{Pros and Cons of Model-Based \ac{kf}-Type Algorithms}
\label{ssec:fund_challneges}
The \ac{kf} and its variants are a family of widely utilized  algorithms~\cite{sarkka2023bayesian}.
The core of these model-based methods is the \ac{ss} model, i.e., the  representation of the system dynamics via closed-form equations, as those in \eqref{eq:ssModel}. When the \ac{ss} model faithfully captures the dynamics, these algorithms have several key desirable properties:
\begin{enumerate}[label={P\arabic*}]
\item \label{itm:Optimality} When the \ac{ss} model is well described as being linear with Gaussian noise, \ac{kf}-based algorithms can approach optimal state estimation performance, in the sense of minimizing the \ac{mse}.
\item \label{itm:Adaptability} Model-based methods are inherently adaptive to known variations in the \ac{ss} model. For instance,  $\myMat{H}_t$ can change with  $t$, and one only needs to substitute the updated matrix in the corresponding equations.
\item \label{itm:interpretability} Their operation is fully interpretable, in the sense that one can associate the internal features with concrete interpretation as they represent, e.g., statistical moments of prior and posterior predictions.
\item \label{itm:uncertainty} They provide reliable uncertainty measures via the error covariance in \eqref{eqn:update2} and \eqref{eq:BW_update2}.  
\item \label{itm:complexity} \ac{kf}-type algorithms operate with relatively low complexity, that does not grow with time. 
\end{enumerate}

However, the reliance of \ac{kf}-type algorithms on faithful mathematical modelling of the underlying dynamics, and their natural suitability with simplistic linear Gaussian models, also gives rise to several core challenges encountered in various applications. These challenges can be roughly categorized as follows:
\begin{enumerate}[label={C\arabic*}]
\item \label{itm:FuncAccuracy} The state evolution and observation models employed in  \ac{ss} models are often {\em approximations of the true dynamics}, whose fidelity may vary considerably between applications. For instance, while the temporal evolution of a state corresponding to the position and velocity of a vehicle can be represented as a linear transformation via  mechanical relationships, e.g., a constant velocity model~\cite{imbiriba2023augmented},
such modeling of $f_t(\cdot)$ is  inherently a  crude first-order approximation. Similarly, the relationship between the velocity of a vehicle and its sensed motor currents can be captured via an observation model of the form \eqref{eq:NLssModel}, the exact specification of the mapping $h_t(\cdot)$ is likely to be elusive.
\item \label{itm:Stochasticity} The purpose of the noise signals $\myVec{v}_t$ and $\myVec{w}_t$ is to $(i)$ capture the inherent stochasticity in the state evolution and measurements, respectively; and $(ii)$ model the discrepancy between the \ac{ss} representation and the true system. Their actual distribution is thus often unknown, complex, and possibly intractable. Non-Gaussianity can have a notable effect on performance and reliability, especially since Kalman-type algorithms seek a linear filtering operation. 
\item \label{itm:NonLinear} Even when the dynamics are faithfully characterized by a non-linear Gaussian \ac{ss} model, Kalman-type algorithms are sub-optimal, with gaps from optimality largely depending on the 
non-linearity. 
\item \label{itm:InferenceSpeed} Despite their relatively low complexity,  non-linear variants of the \ac{kf}, such as the \ac{ekf} and \ac{ukf}, induce some latency during filtering. This is due to the need to carry out, e.g., local linearization and matrix inversion, on each time instant (which the linear \ac{kf} can do offline based on knowledge of the statistics). 
\end{enumerate}

\textcolor{NewColor}{The above characterization of the desirable properties and the challenges of model-based \ac{kf}-type algorithms is summarized in Table~\ref{tab:myProsConsSummary}. Specifically, the challenges characterized in \ref{itm:FuncAccuracy}-\ref{itm:InferenceSpeed}} 
motivate exploring data-driven approaches for tackling tasks associated with \ac{ss} models, as detailed in the following section.

\begin{table}
\centering 
\setlength{\tabcolsep}{2pt} 
\renewcommand{\arraystretch}{1.5}
{\scriptsize
\color{NewColor}
    \begin{tabular}{|p{0.5cm} p{5.2cm}| p{0.5cm} p{5.2cm}|}
    \hline
       \multicolumn{2}{|c|}{{\bf Desirable Properties}} &  \multicolumn{2}{|c|}{{\bf Challenges}}\\
       \hline
       \hline
       \ref{itm:Optimality}  & Optimal for linear Gaussian  models & \ref{itm:FuncAccuracy} & \ac{ss} model is typically approximated \\
       \ref{itm:Adaptability} & Adaptable to known variations & \ref{itm:Stochasticity} & Noise model is often elusive \\
       \ref{itm:interpretability} & Interpretable operation & 
       \ref{itm:NonLinear} & Sub-optimal for non-linear  models \\
       \ref{itm:uncertainty} & Provide uncertainty &
       \ref{itm:InferenceSpeed} &Latency for non-linear  models \\
       \ref{itm:complexity} & Relatively low complexity & & \\
       \hline
    \end{tabular}
}
    \caption{Summary of desired properties and challenges of model-based \ac{kf}-type algorithms.}
    \label{tab:myProsConsSummary}
    \color{black}
\end{table}

\section{Combining AI with KFs} 
\label{sec:combining}
Recent years have witnessed remarkable  success of deep learning, being the main enabler  for \ac{ai},   in various applications involving processing of time sequences. Data-driven \acp{dnn}  were shown to be able to catch the subtleties of complex processes and replace the need to explicitly characterize the domain of interest. Therefore, an alternative strategy to implement state estimation while coping with \ref{itm:FuncAccuracy}-\ref{itm:InferenceSpeed}, namely, without requiring explicit and accurate knowledge of the \ac{ss} representation, is to learn this task from data using deep learning.

\subsection{Time Sequence Filtering with \acp{dnn}}
\label{ssec:FilteringDNNs}
A common \ac{ml} strategy utilizes highly parameterized abstract models  trained from data to find the parameterization that minimizes the empirical risk (with regularization introduced to prevent overfitting). Their mapping, denoted  $\dnnFunc$, is dictated by a set of parameters denoted $\dnnParam$. In deep learning, the parametric model $\dnnFunc$ is a \ac{dnn}, with $\dnnParam$ being the network parameters.  Such highly-parametrized abstract models can effectively approximate any Borel measurable mapping, as follows from the universal approximation theorem \cite[Ch. 6.4.1]{goodfellow2016deep}.

\acp{dnn} can learn various tasks from data without requiring  mathematical representation of the underlying dynamics and observations model. Accordingly, \acp{dnn} applied for tasks such as filtering and smoothing do not require the formulation of the dynamics as a \ac{ss} representation. Despite this invariance, the resulting architecture of the \ac{dnn} can still draw some inspiration from \ac{ss} representation and \ac{kf}-type processing. Consequently, we divide our presentation to {\em conventional \acp{dnn}}, which only account for the fact that the data being processed is a time sequence, and {\em \ac{ss}/\ac{kf}-inspired \acp{dnn}}, with both families being invariant of the underlying model.

\subsubsection{Conventional \acp{dnn}}
\label{sssec:standard_dnns}
Tasks associated with \ac{ss} models, particularly filtering and smoothing, involve processing  time sequences with temporal correlations. 
Accordingly, \acp{dnn} designed for time sequences can conceptually be trained to carry out these tasks. These architectures, discussed next, somewhat deviate from the basic \ac{dnn} form, i.e., \ac{fc} layers, which process fixed-size vectors rather than time sequences.

\begin{figure}
    \centering
    \includegraphics[width=\figWidth]{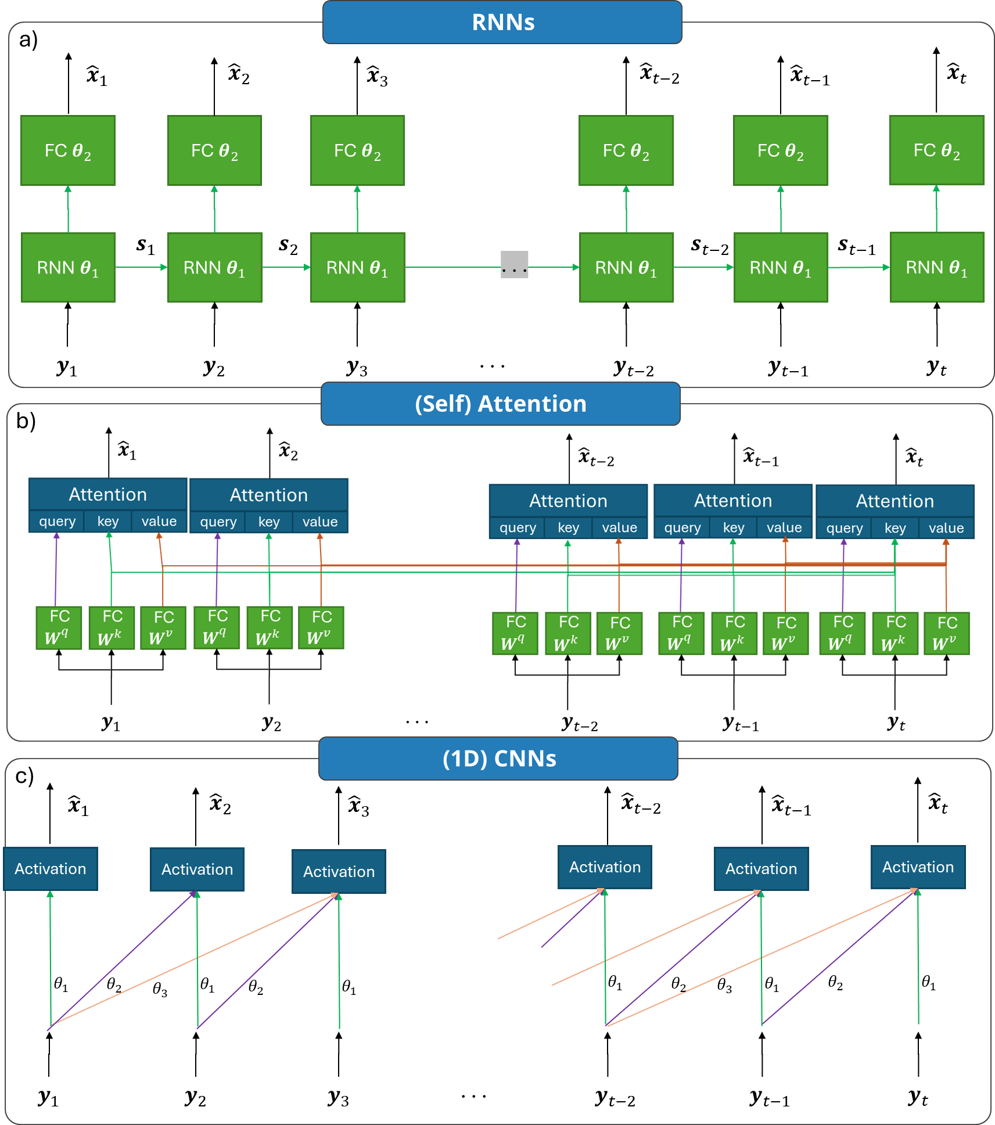}
    \caption{Illustration of conventional \ac{dnn} architectures for tasks related to filtering and smoothing, including $(a)$ \acp{rnn}; $(b)$ Attention; and $(c)$ \acp{cnn}.}
    \label{fig:ConventionalDNNs}
\end{figure}

{\bf RNNs:} 
A common \ac{dnn} architecture  for time sequences is based on \acp{rnn}. \acp{rnn} maintain an internal state vector, denoted $\myVec{s}_t$, representing the system memory~\cite[Ch. 10]{goodfellow2016deep}. The parametric mapping is given by:
\begin{equation}
\label{eqn:RNNb}
\hat{\myVec{x}}_t= \dnnFunc(\myVec{y}_t, \myVec{s}_{t-1}),
\end{equation}
where the internal state also evolves via a learned parametric mapping 
\begin{equation}
\label{eqn:RNNa}
\myVec{s}_{t} = \myFunc{G}_{\dnnParam}(\myVec{y}_t, \myVec{s}_{t-1}).
\end{equation}
 The vanilla implementation of an \ac{rnn} uses a hidden layer to map $\myVec{y}_t$ and the latent $\myVec{s}_{t-1}$  into an updated hidden variable $\myVec{s}_t$. Alternative \ac{rnn} architectures, such as \acp{gru} and \acp{lstm}, employ several learned cells to update $\myVec{s}_t$. 
 Then, $\myVec{s}_t$ is used to generate the instantaneous output  $\hat{\myVec{x}}_t$ using another layer, as illustrated in Fig.~\ref{fig:ConventionalDNNs}(a). 

{\bf Attention:} 
A widely popular \ac{dnn} architecture, which is the core of the transformer model, is based on attention mechanisms~\cite{vaswani2017attention}. Such architectures were shown to be extremely successful in learning complex tasks in natural language processing, where the considered signals can be viewed as time sequences. 

\color{NewColor}
Attention-based \acp{dnn} are based on the mathematical representation of the {\em attention mechanism}, which is a simple mathematical formulation of how the human brain divides its attention. It is a function of $n_a$ key-value pairs $\{\myVec{k}_i,\myVec{\nu}_i\}_{i=1}^{n_a}$, representing the environment being sensed, and a query vector $\myVec{q}$. 
The most common form of the attention mechanism is that of {\em scaled dot-product attention}~\cite{vaswani2017attention}, which  processes $\myVec{q}$ and $\{\myVec{k}_i,\myVec{\nu}_i\}_{i=1}^{n_a}$ into a vector given by  
\begin{equation*}
    {\rm Attention}\left(\myVec{q},\{\myVec{k}_i,\myVec{\nu}_i\}_{i=1}^{n_a}\right)=\sum_{i=1}^{n_a}  {\rm softmax}\big({\rm const} \cdot \myVec{q}^T\myVec{k}_i\big) \myVec{\nu}_i.      
\end{equation*}

When applied for tasks such as filtering or smoothing, the attention mechanism is used as a trainable layer as a form of {\em self-attention},  illustrated in Fig.~\ref{fig:ConventionalDNNs}(b). A self-attention head is an \ac{ml} model that applies trained linear layers to map the input sequence in order to obtain  the queries, keys, and values, processed via an attention mechanism. A single head self-attention with parameters $\dnnParam=\{\myMat{W}^q, \myMat{W}^k, \myMat{W}^v\}$ applied for smoothing can be written as 
$ \hat{\myVec{x}}_{1:T} = \dnnFunc\left(\myVec{y}_{1:T}\right)$, where
\begin{align}
    \hat{\myVec{x}}_t 
    &= \sum_{\tau=1}^{T}  {\rm softmax}\left( {\rm const} \cdot (\myMat{W}^q\myVec{y}_t)^T(\myMat{W}^k\myVec{y}_\tau)\right) \myMat{W}^v\myVec{y}_\tau.
    \label{eqn:Attention}
\end{align}
\color{black} 

Attention-based \acp{dnn} processing time sequences typically apply multiple mappings as in \eqref{eqn:Attention} in parallel, as a form of {\em multi-head attention}. As opposed to \acp{rnn}, attention mechanisms do not maintain an internal state vector that is sequentially updated. This makes attention-based \acp{dnn} more amenable to parallel training compared with \acp{rnn}. However, attention mechanisms as in \eqref{eqn:Attention} are invariant of the order of the processed signal samples, and are thus typically combined with additional pre-processing termed {\em positional embedding} that embed each sample while accounting for its position.  

{\bf CNNs:} 
Unlike \acp{rnn} and attention, \acp{cnn} are  architectures that originate from image processing, and not time sequences. Specifically, \acp{cnn} are designed to learn the parameters of spatial kernels, aiming to exploit the locality and spatial stationarity of image data~\cite[Ch. 9]{goodfellow2016deep}. Nonetheless, \acp{cnn} can also be applied for time sequence processing, and particularly for tasks such as filtering and smoothing. 
For once, the trainable kernel of \acp{cnn} can implement a learned finite impulse response filter (as a form of 1D \ac{cnn}) and be applied to a time sequence, as illustrated in Fig.~\ref{fig:ConventionalDNNs}(c). Alternatively, one can apply \acp{cnn} to the time sequence by first converting the time sequence (or an observed window) into the form of an image via, e.g., short-time Fourier transform, and then use a \ac{cnn} to process this representation.

\begin{tcolorbox}[float*=t,
    width=\boxWidth,
	toprule = 0mm,
	bottomrule = 0mm,
	leftrule = 0mm,
	rightrule = 0mm,
	arc = 0mm,
	colframe = myblue,
	colback = mypurple,
	fonttitle = \sffamily\bfseries\large,
	title = Selective State-Space Models\acused{ssm} (\acp{ssm})]	
	\label{Box:mamba}
\begin{wrapfigure}{r}{5cm} 
    \centering
    \includegraphics[width=5cm]{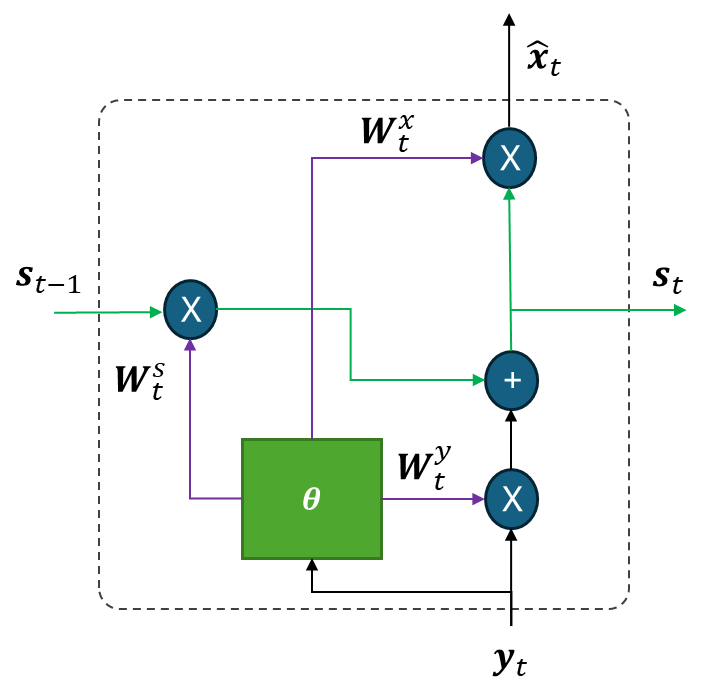} 
    \caption{Selective \ac{ssm}.} 
    \label{fig:SelectiveSSM1}
\end{wrapfigure}
An emerging \ac{ssm} \ac{dnn}, which is the core of the Mamba architecture and its variants~\cite{gu2023mamba}, is the {\em selective \ac{ssm}}~\cite{gu2021efficiently}. Selective \acp{ssm} are \ac{ml} models that process time sequences via \eqref{eqn:SSM}. However, instead of using fixed linear mappings as in \eqref{eqn:SSM}, here  $\myMat{W}^y, \myMat{W}^s, \myMat{W}^x$ are obtained from learned mappings of the input $\myVec{y}_t$, effectively implementing a time-varying and non-linear \ac{ssm}.

Generally speaking, consider the discretization with sampling period $\Delta$ of continuous-time deterministic \ac{ss} representations. The resulting state evolution in \eqref{eqn:SSMa} is parameterized by 
\begin{subequations}
    \label{eqn:Mamba}
\begin{align}
\label{eqn:MambaA}
    \myMat{W}^s &= \exp(\Delta \cdot \bar{\myMat{W}}^s) \\
    \myMat{W}^y &= (\Delta \cdot \bar{\myMat{W}}^s)^{-1}(\exp(\Delta \cdot \bar{\myMat{W}}^s) - \myMat{I})  \Delta \cdot \bar{\myMat{W}}^y,
    \label{eqn:MambaB}
\end{align}
\end{subequations}
where $\bar{\myMat{W}}^y,\bar{ \myMat{W}}^s$ are the continuous-time state evolution matrices. Selective \acp{ssm} implement \acp{ssm} based on this representation, while using a dedicated layer with parameters $\dnnParam_1$ to map $\myVec{y}_t$ into $\bar{\myMat{W}}^y$, $\Delta$, and $\myMat{W}^x$, where the former two are then substituted in \eqref{eqn:Mamba} along with the learned $ \bar{\myMat{W}}^s$. The overall parameters are thus $\dnnParam=\{\dnnParam_1, \bar{\myMat{W}}^s\}$. The resulting operation is illustrated in Fig.~\ref{fig:SelectiveSSM1}. 
\end{tcolorbox}

\subsubsection{\ac{ss}/\ac{kf}-Inspired \acp{dnn}}
\acp{dnn} applied to process time sequences, are invariant of the underlying statistics governing the dynamics, and learn their operation purely from data. Nonetheless, one can still design \acp{dnn} for processing time sequences whose architecture is to some extent inspired by traditional model-based processing of such signals, particularly on \ac{ss} representations and \ac{kf} processing. 

{\bf \ac{ss}-Inspired Architectures:}
A form of \ac{dnn} architecture that is inspired by \ac{ss} representation models the filter (and not the dynamics as in \eqref{eq:ssModel}), as a {\em deterministic \ac{ss} model}. These \ac{ml} models, which we term \ac{ssm} following~\cite{gu2021combining}, parameterize the mapping from $\myVec{y}_t$ into the estimate $\hat{\myVec{x}}_t$ using a latent state vector $\myVec{s}_{t-1}$. Particularly, the architecture operates using two linear mappings, where first $\myVec{s}_t$ is updated via
\begin{subequations}
    \label{eqn:SSM}
\begin{equation}
\label{eqn:SSMa}
    \myVec{s}_t = \myMat{W}^y \myVec{y}_t + \myMat{W}^s\myVec{s}_{t-1}.
\end{equation}
Then, the estimate is given by
\begin{equation}
\label{eqn:SSMb}
    \hat{\myVec{x}}_t = \myMat{W}^x \myVec{s}_t.
\end{equation}
\end{subequations}
The parameters of the resulting \ac{ml} architecture are $\dnnParam = \{\myMat{W}^y, \myMat{W}^s, \myMat{W}^x\}$.

The \ac{ssm} operation in \eqref{eqn:SSM} can be viewed as a form of an \ac{rnn}, with \eqref{eqn:SSMa} implementing \eqref{eqn:RNNa}, and \eqref{eqn:SSMb} implementing \eqref{eqn:RNNb}. While the restriction to linear time-invariant learned operators in \eqref{eqn:SSM} often results in limited effectiveness, an extension of \acp{ssm}, termed {\em Selective \acp{ssm}} (see box on page~\pageref{Box:mamba}), was recently shown to yield efficient and high-performance architectures that are competitive to costly and complex transformers~\cite{gu2023mamba}.

{\bf \ac{kf}-Inspired Architectures:}
While \acp{ssm} parameterize a learned filter via a \ac{ss} representation, one can also design \acp{dnn}  architectures inspired by filters suitable for tracking in \ac{ss} models, e.g., \ac{kf}-type algorithms.  An example of such architectures is the \ac{rkn}, proposed in \cite{becker2019recurrent}. 

The \ac{rkn} is a \ac{dnn} whose operation imitates the prediction-update steps of the \ac{kf} and the \ac{ekf}. However, while the model-based algorithms realize these computations based on the characterization of the dynamics as a \ac{ss} model, the \ac{rkn} parameterizes the predict and update stages as two interconnected \acp{dnn}. The predict \ac{dnn}, whose parameters are $\dnnParam_1$, replaces \eqref{eqn:predict_evol}, and computes the a-priori state prediction via
\begin{subequations}
    \label{eqn:RKN}
\begin{equation}
    \hat{\myVec{x}}_{t\given{t-1}}, \myMat{\Sigma}_{t\given{t-1}} = \myFunc{F}_{\dnnParam_1}\left(\hat{\myVec{x}}_{t-1\given{t-1}}, \myMat{\Sigma}_{t-1\given{t-1}}\right).
\end{equation}
Similarly, the update \ac{dnn}, parameterized by $\dnnParam_2$, produces the posterior state estimate of \eqref{eqn:update} via 
\begin{equation}
    \hat{\myVec{x}}_{t\given{t}}, \myMat{\Sigma}_{t\given{t}} = \myFunc{G}_{\dnnParam_2}\left(\hat{\myVec{x}}_{t\given{t-1}}, \myMat{\Sigma}_{t\given{t-1}}, \myVec{y}_t\right).
\end{equation}
\end{subequations}
The resulting architecture can be combined with additional learned input and output processing \cite{becker2019recurrent}.

\subsubsection{Pros and Cons of \ac{dnn}-Based State Estimation}
The \ac{dnn} architectures detailed so far can be trained to carry out state estimation tasks, i.e., map the observed time sequence into the latent state sequence.  Specifically, provided data describing the task, e.g., labeled data set comprised of trajectories of observations and corresponding states, these \acp{dnn} learn their mapping from data without relying on any statistical modeling of the dynamics. This form of {\em discriminative learning}, i.e., leveraging data to learn to carry out a task end-to-end~\cite{shlezinger2022discriminative}, excels where model-based methods struggle: it is not affected by inaccurate modeling (thus coping with \ref{itm:FuncAccuracy}-\ref{itm:Stochasticity}, and their abstractness allows \acp{dnn} to operate reliably in complex settings (\ref{itm:NonLinear}). In terms of inference speed (\ref{itm:InferenceSpeed}), while \acp{dnn} involve lengthy and complex training, they often provide rapid inference (forward path), particularly when using relatively compact parameterization. This follows as trained \acp{dnn} are highly amenable to parallelization and acceleration of built-in hardware software accelerators, e.g., PyTorch.

Nonetheless,  replacing \ac{kf}-type algorithms with \acp{dnn} trained end-to-end gives rise to various shortcomings, particularly in losing some of the desired properties of model-based methods. For once, \acp{dnn} lack in adaptability (\ref{itm:Adaptability}), as one cannot substitute time-varying parameters of \ac{ss} models into their operation, and thus changes may necessitate lengthy retraining. 
Moreover, \acp{dnn} are typically highly parameterized architectures viewed as black-boxes, that do not share the interpretability of model-based \ac{kf}-type methods (\ref{itm:interpretability}), and are complex to train or even store on limited devices (\ref{itm:complexity}).
Furthermore, \acp{dnn} struggle in providing uncertainty (\ref{itm:uncertainty}), for which there is typically no "ground truth" to learn from, and do not share the theoretical guarantees of \acp{kf} (\ref{itm:Optimality}).

\subsection{\ac{ai}-Augmented \acp{kf}}
\label{ssec:combining_kfs}
 
The \ac{dnn}-based approaches discussed so far are highly data-driven, in the sense that they do not rely on any statistical characterization of the dynamics. Even \ac{ss} or \ac{kf} inspired architectures, such as the \ac{rkn} whose operation generally follows the high-level stages of the \ac{kf}, are ignorant of any \ac{ss} modeling. 

An alternative approach that aims to benefit from the best of both worlds employs hybrid model-based/data-driven designs via model-based deep learning
~\cite{shlezinger2020model,shlezinger2022model}.
Such algorithms typically jointly leverage data along with some form of domain knowledge, i.e., partial knowledge of some components of the underlying \ac{ss} model (which is often available to some degree as noted in \ref{itm:FuncAccuracy}). For state estimation,  hybrid algorithms  {\em augment} \ac{kf}-type algorithms with deep learning modules, rather than replacing them with \acp{dnn}.

\begin{figure*}
    \centering
    \includegraphics[width=\linewidth]{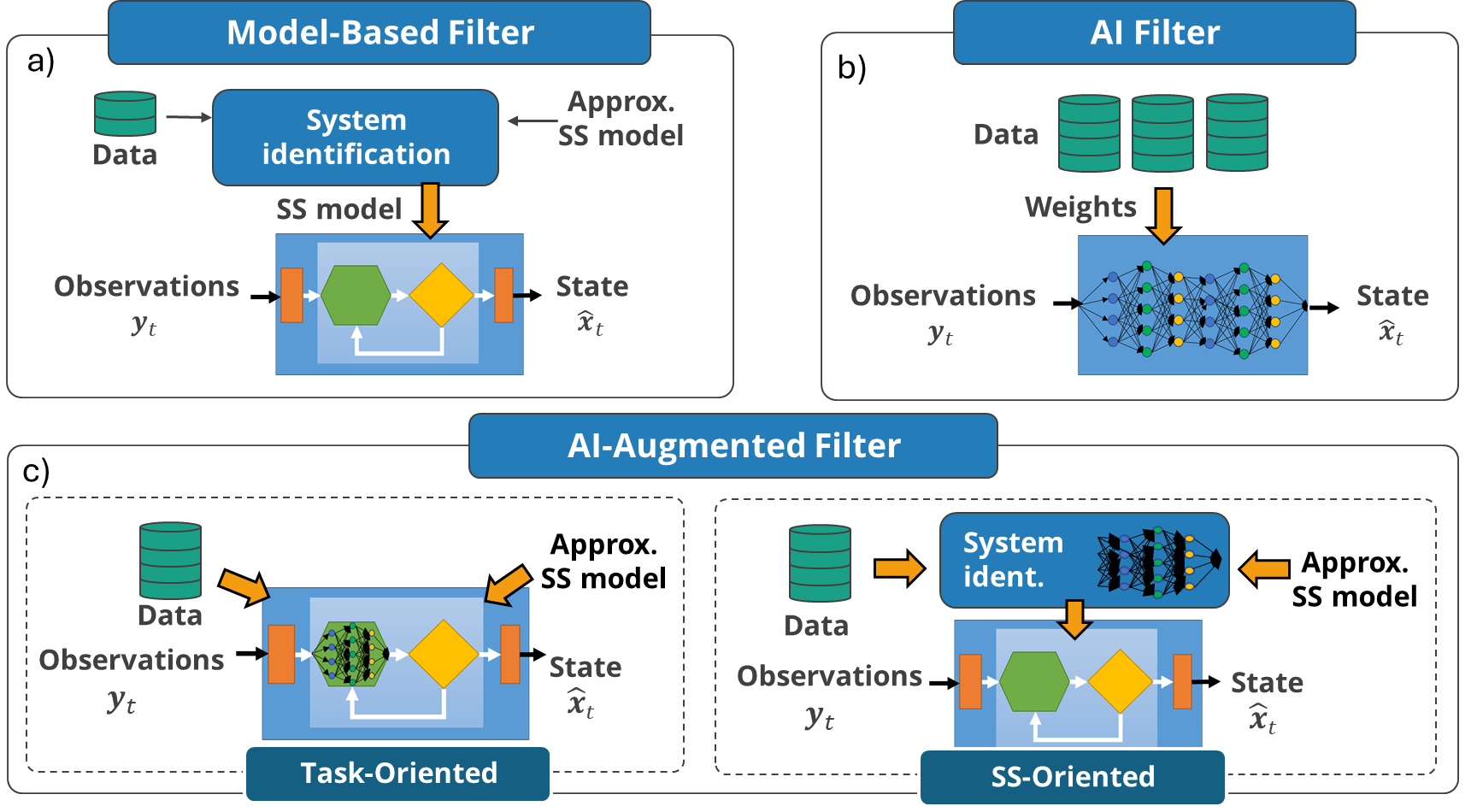}
    \caption{Illustrative comparison between model-based Kalman-type filters ($a$); \ac{ai}-based filters ($b$); and \ac{ai}-augmented \ac{kf} ($c$) divided into task-oriented and \ac{ss}-oriented designs. }
    \label{fig:Comparison1}
\end{figure*}

As discussed above, the direct application of \acp{dnn} 
reviewed so far is based on task-oriented end-to-end discriminative learning. In the same spirit, existing approaches   for augmenting  \acp{kf} with \ac{ai} tools can be generally categorized following the \ac{ml} paradigms of generative and discriminative learning~\cite{shlezinger2022discriminative}: 
\begin{itemize}
    \item {\em \ac{ss}-oriented} hybrid algorithms, that use data to learn the underlying statistical model as \ac{dnn}-aided system identification (thus bearing similarity to generative learning).
    \item  {\em Task-oriented} schemes, that directly learn to carry out the state estimation task (as a form of discriminative learning), while leveraging partial state knowledge and principled \ac{kf} stages as inductive bias. 
\end{itemize}
This categorization, illustrated in Fig.~\ref{fig:Comparison1}, serves for our  review of existing approaches in the sequel.

\section{\ac{ai}-Augmented \acp{kf} via Task-Oriented Learning}
\label{sec:discriminative}
The first family of \ac{ai}-aided \acp{kf} converts Kalman-type state estimation into an \ac{ml} architecture that is trainable end-to-end via \ac{dnn}-augmentation. The key rationale is to leverage deep learning techniques to directly learn the state estimation as a form of discriminative learning. A common approach to designing such architectures is based on using an external \ac{dnn}, operating alongside a classic state estimator, with recent architectures integrating deep learning modules into the internal processing of Kalman-type algorithms.

\subsection{External \ac{dnn} Architectures}
\label{ssec:discriminative_external}
Utilizing external \acp{dnn} aims to enhance \ac{kf}-type algorithms without altering their internal processing. This approach facilitates design, as one can separate the \ac{dnn} components from the classic state estimator. Broadly speaking, the leading approaches to utilizing external \acp{dnn} employ them either sequentially, i.e., for pre-processing, or in parallel, e.g., as learned correction terms.

\subsubsection{Learned Pre-Processing}
A popular approach when dealing with complex measurement models, e.g., visual or multi-modal observations, builds on the ability of \acp{dnn} to extract meaningful features from complex data. Specifically, it uses a  \ac{dnn} pre-processor to map the observations into a latent space~\cite{zhou2020kfnet, coskun2017long}.

\begin{figure}
    \centering
    \includegraphics[width=\figWidth]{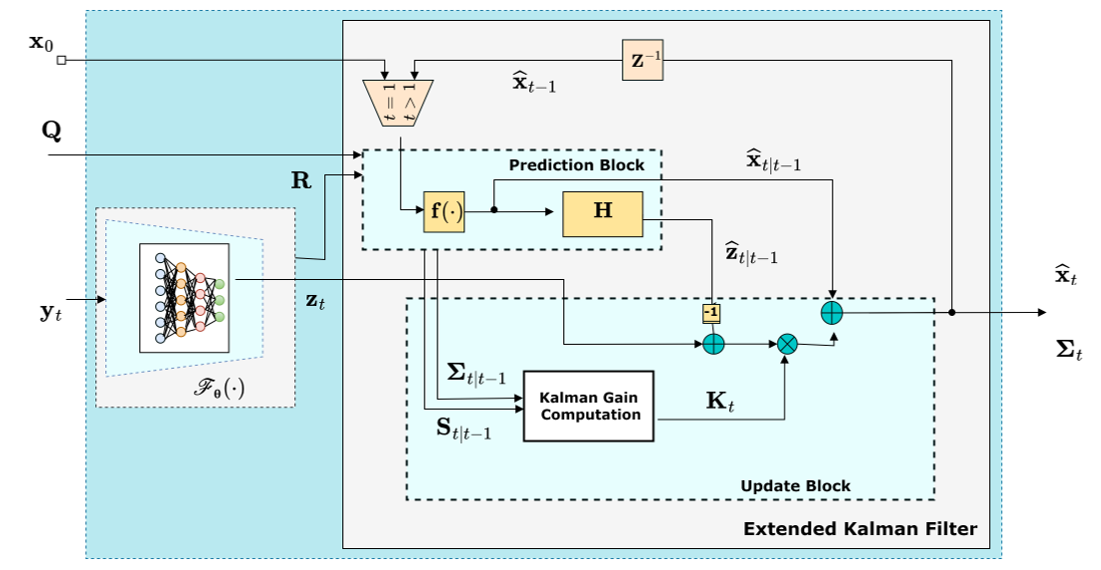}
    \caption{\ac{ekf} with \ac{dnn} pre-processing illustration.}
    \label{fig:SeperateDNN}
\end{figure}

{\bf Architecture:}
Consider a \ac{ss} model where the observation model $h_t(\cdot)$ is complex and possibly intractable. In such cases, one can design a \ac{dnn} with parameters $\dnnParam$ whose output is approximated as obeying a linear Gaussian observations model, i.e.,
\begin{equation}
\label{eqn:latent}
    \myVec{z}_t = \dnnFunc(\myVec{y}_t) \approx \myMat{H} \myVec{x}_t + \myVec{w}_t, \qquad \myVec{w}_t \sim \mathcal{N}(\myVec{0}, \myMat{R}),
\end{equation}
with the observation matrix $\myMat{H}$ being fixed a-priori.  
The latent $\myVec{z}_t$, combined with a known state evolution model, represent observations of a closed-form \ac{ss} representation. Accordingly, the latent signal can be used as input to a model-based \ac{kf}-type state estimator to track $\myVec{x}_t$, as illustrated in Fig.~\ref{fig:SeperateDNN}. 

{\bf Supervised Training:} 
In a supervised setting, one has access to a labeled data set comprised of multiple pairs of state and observation length $T$ trajectories, e.g., 
\begin{equation}
    \label{eqn:DataSet}
    \mySet{D}_{\rm s} = \big\{\myVec{x}_{1:T}^{(i)}, \myVec{y}_{1:T}^{(i)}\big\}_{i=1}^{\Ntraining}.
\end{equation}
Given such  data,  $\dnnParam$ can be trained  by encouraging the \ac{dnn} output to closely match $\myMat{H} \myVec{x}_t$, i.e., using a loss 
\begin{equation}
    \label{eqn:lossSeperate1}
    \mathcal{L}_{\mySet{D}_{\rm s}}(\dnnParam) = \frac{1}{\Ntraining T}\sum_{i=1}^{\Ntraining}\sum_{t=1}^{T} \left\|\dnnFunc(\myVec{y}_t^{(i)}) - \myMat{H} \myVec{x}_t^{(i)}\right\|_2.
\end{equation}
When training via \eqref{eqn:lossSeperate1}, the noise covariance $\myMat{R}$ can be estimated empirically from the validation data. 

An alternative training approach finds $\dnnParam$ based on its downstream state estimation task, leveraging the differentiable operation of \acp{kf} \cite{xu2021ekfnet}. 
\color{NewColor}
Here, by letting $\hat{\myVec{x}}_t\big(\dnnFunc(\myVec{y}_{1:t})\big)$ be the state estimate obtained by a \ac{kf}/\ac{ekf} with observations sequence $\myVec{z}_{1:t}$, where $\myVec{z}_\tau = \dnnFunc(\myVec{y}_\tau)$, a candidate loss measure is 
\begin{equation}
    \label{eqn:lossJoint1}
    \mathcal{L}_{\mySet{D}_{\rm s}}(\dnnParam) = \frac{1}{\Ntraining T}\sum_{i=1}^{\Ntraining}\sum_{t=1}^{T} \left\|\hat{\myVec{x}}_t\big(\dnnFunc(\myVec{y}_{1:t}^{(i)})\big) -  \myVec{x}_t^{(i)}\right\|_2.
\end{equation}
\color{black}
The covariance  $\myMat{R}$ can be learned during training by being incorporated into the trainable parameters.

{\bf Discussion:} 
Using an external pre-processing \ac{dnn} that is separated from the Kalman-type state estimation algorithm is a design approach geared mostly towards handling complex observation models. As state estimation is carried out by a model-based algorithm, it preserves most of its favorable properties: 
The state estimation operation is interpretable (\ref{itm:interpretability}), and uncertainty measures are provided (\ref{itm:uncertainty}). The excessive complexity lies in the incorporation of the pre-processing \ac{dnn} and thus depends on its parameterization, as well as on the dimensions of $\myVec{z}_t$. For instance, when $\myVec{z}_t$ is of a much lower dimension compared to $\myVec{y}_t$, the complexity and inference latency savings of applying a \ac{kf} to $\myVec{z}_t$ compared to using $\myVec{y}_t$ as observations may surpass the added complexity and latency of the pre-processing \ac{dnn}. 

The adaptivity of \acp{kf} (\ref{itm:Adaptability}) is not necessarily preserved. Specifically, when the \ac{dnn} is fully separate from the state estimator (e.g., when training via~\eqref{eqn:lossSeperate1}), then the state evolution parameters only affect the model-based algorithm, and thus one can still operate with time-varying ${f}_t(\cdot)$ (assuming its variations are known). This is not necessarily the case when training via \eqref{eqn:lossJoint1}, as the latent features are learned to be ones most supporting state estimation based on the evolution model used during training~\cite{buchnik2023latent}. Moreover, variations in the observation model typically necessitate re-training of $\dnnParam$. 

Using an external pre-processing \ac{dnn}, while being simple and straightforward to combine with model-based state estimation,  relies on Gaussianity and known distribution of the state evolution. The resulting latent \ac{ss} model is often non-Gaussian, which can impact the tracking accuracy in the latent space, thus not handling \ref{itm:Stochasticity}. Moreover, this approach does not cope with complexities in the state evolution, thus not being geared towards handling inaccuracies (\ref{itm:FuncAccuracy}) and dominant non-linearities (\ref{itm:NonLinear}) in ${f}_t(\cdot)$.

\subsubsection{Learned Correction Terms}

In scenarios with full characterization of the dynamic system as a \ac{ss} model, such that \ac{kf}-type algorithms are applicable, yet the characterization is not fully accurate, external \acp{dnn} can provide correction terms to internal computations of the model-based method
~\cite{satorras2019combining}. This is illustrated using a representative example, based on the augmented Kalman smoother proposed in \cite{satorras2019combining}. 

{\bf Architecture:} 
Consider a linear Gaussian \ac{ss} model, and focus on the smoothing task, i.e., recovering a sequence of $T$ state variables 
$\myVec{x}_{1:T}$  from the entire observed sequence 
$\myVec{y}_{1:T}$. As discussed when presenting the \ac{rts} algorithm, various algorithms exist for such tasks, one of which involves smoothing by seeking to maximize the log-likelihood function via gradient ascent optimization, i.e., by iterating over 
\begin{equation}
   \label{eqn:GradKal}
    \myVec{x}^{(q+1)}_{1:T} = \myVec{x}^{(q)}_{1:T} + \gamma \nabla_{\myVec{x}^{(q)}_{1:T}}\log \pdf\left(\myVec{y}_{1:T}, \myVec{x}_{1:T}^{(q)} \right),
\end{equation}
where $\gamma>0$ is a step-size, for $q=0,1,2,\ldots$ denoting the iteration index.  The Markovian nature of the \ac{ss} model indicates that the gradients in \eqref{eqn:GradKal} can be computed via message passing, such that for the $t$'th index
\begin{equation}
    \nabla_{\myVec{x}_t^{(q)}}\log \pdf\left(\myVec{y}_{1:T}, \myVec{x}^{(q)}_{1:T}\right) 
     =\mu^{(q)}_{\myVec{x}_{t-1}\rightarrow \myVec{x}_t} +  \mu^{(q)}_{\myVec{x}_{t+1}\rightarrow \myVec{x}_t} + \mu^{(q)}_{\myVec{y}_{t}\rightarrow \myVec{x}_t}, 
 \end{equation}
where the summands, referred to as messages, are given by 
\begin{subequations}
    \label{eqn:KalSmoothQuant1}
    \begin{align}
    \mu^{(q)}_{\myVec{x}_{t-1}\rightarrow \myVec{x}_t} & =  -\myMat{Q}^{-1}\left( \myVec{x}_t^{(q)} - \myMat{F} \myVec{x}_{t-1}^{(q)}  \right),  \\
    \mu^{(q)}_{\myVec{x}_{t+1}\rightarrow \myVec{x}_t} & =  \myMat{F}^T\myMat{Q}^{-1}\left( \myVec{x}_{t+1}^{(q)} - \myMat{F} \myVec{x}_{t}^{(q)}  \right),  \\
    \mu^{(q)}_{\myVec{y}_{t}\rightarrow \myVec{x}_t} & =  \myMat{H}^T\myMat{R}^{-1}\left( \myVec{y}_{t}- \myMat{H} \myVec{x}_{t}^{(q)}  \right).
    \end{align}
\end{subequations} 
The iterative procedure in \eqref{eqn:GradKal}, is repeated until convergence, and the resulting $\myVec{x}_{1:T}^{(q)}$ is used as the estimate. In \eqref{eqn:KalSmoothQuant1}, the messages are obtained by assuming a time-invariant version of the SS model in \eqref{eq:LinssModel} \cite{satorras2019combining}. 

 An external \ac{dnn} augmentation suggested in \cite{satorras2019combining} enhances the smoother under approximated \ac{ss} characterization by learning to map the messages in \eqref{eqn:KalSmoothQuant1} into a correction term $\myVec{\epsilon}^{(q+1)}_{1:T}$, replacing the update rule \eqref{eqn:GradKal} with 
\begin{equation}
\label{eqn:GradKal2}
\myVec{x}^{(q+1)}_{1:T} = \myVec{x}^{(q)}_{1:T} + \gamma \left(\nabla_{\myVec{x}_{1:T}^{(q)}}\log \pdf\left(\myVec{y}_{1:T}, \myVec{x}^{(q)}_{1:T}\right) + \myVec{\epsilon}^{(q+1)}_{1:T} \right).
\end{equation}
    Particularly, based on the representation of the smoothing operation as messages exchanged over a Markovian factor graph whose nodes are the state variables,  \cite{satorras2019combining} proposed a \ac{gnn}-\ac{rnn} architecture with parameters $\dnnParam$. This architecture maps the messages in \eqref{eqn:KalSmoothQuant1} into the correction term $\myVec{\epsilon}^{(q+1)}_{1:T}$, namely, 
    \begin{equation}
        \myVec{\epsilon}^{(q+1)}_{1:T} = \dnnFunc\left(\myVec{y}_{1:T}, \{\mu^{(q)}_{\myVec{x}_{t-1}\rightarrow \myVec{x}_t}, \mu^{(q)}_{\myVec{x}_{t+1}\rightarrow \myVec{x}_t},  \mu^{(q)}_{\myVec{y}_{t}\rightarrow \myVec{x}_t}\}_{t=1}^T \right). 
        \label{eqn:CorrectTerm}
    \end{equation}

{\bf Supervised Training:} 
The external \ac{dnn} correction mechanism is trained end-to-end, such that the state predicted by the corrected algorithm best matches the true state. Since smoothing here is based on an iterative algorithm (gradient ascent over the likelihood objective), its training is carried out via a form of deep unfolding~\cite{shlezinger2022model}, fixing the number of iterations to some $Q$. Then, as  the iterative steps in \eqref{eqn:GradKal2} should provide gradually refined state estimates, the loss function accounts for the contribution of the intermediate iterations with a monotonically increasing contribution. Particularly, the loss function used for training $\dnnParam$ is 
\begin{equation}
    \label{eqn:lossExternal}
    \mathcal{L}_{\mySet{D}_{\rm s}}(\dnnParam) = \frac{1}{\Ntraining T}\sum_{i=1}^{\Ntraining}\sum_{t=1}^{T} \sum_{q=1}^Q\frac{q}{Q}\left\|\hat{\myVec{x}}_t^{(q)}\big(\myVec{y}^{(i)}_{1:T}; \dnnParam\big) -  \myVec{x}_t^{(i)}\right\|_2,
\end{equation}
 where $\hat{\myVec{x}}_t^{(q)}(\myVec{y}_{1:T}^{(i)}; \dnnParam)$ is the smoothed estimate of $\myVec{x}_t^{(i)}$ produced by the $q$'th iteration, i.e., via \eqref{eqn:GradKal2}. 

{\bf Discussion:} 
Augmenting a \ac{kf}-type algorithm via a learned correction term is useful when one can still apply the model-based algorithm quite reliably (up to some possible errors that external \ac{dnn} corrects). For instance, the augmented algorithm  above approximates the dynamic system  as a linear Gaussian \ac{ss} model, such that model-based smoothing can be applied based on this formulation, yet it is not \ac{mse} optimal (as is the case without discrepancy in the \ac{ss} model). Consequently, this approach is suitable for tackling \ref{itm:FuncAccuracy}, yet it is less valid for  complex dynamics (\ref{itm:NonLinear}), and only adds excessive latency to the model-based algorithm (\ref{itm:InferenceSpeed}). 

\subsection{Integrated \ac{dnn} Architectures}
\label{ssec:discriminative_integrated}
Unlike external architectures, which employ \acp{dnn} separately from Kalman-type algorithms, integrated architectures replace intermediate computations with \acp{dnn}. Doing so converts the algorithm into a trainable \ac{ml} model which follows the operation of the classic  method as an inductive bias. The key design rationale is to augment computations that depend on missing domain knowledge with dedicated \acp{dnn}. Accordingly, existing designs vary based on the absent domain knowledge or, alternatively, on the augmented computation. 

\subsubsection{Learned Kalman Gain}
\label{ssec:discriminative_ss}
 A key part of Kalman-type algorithms is the derivation of the \ac{kg} $\Kgain_t$. In particular, its computation via \eqref{eq:FWGain} encapsulates the need to propagate the second-order moments of the state and observations. This, in turn, induces the requirement to have full knowledge of the underlying stochasticity (\ref{itm:Stochasticity}); leads to some of the core challenges in dealing with non-linear \ac{ss} models (\ref{itm:NonLinear}); and results in excessive latency which is associated with propagating these moments (\ref{itm:InferenceSpeed}). Accordingly, a candidate approach for state estimation in partially known \ac{ss} models, which is the basis for the KalmanNet algorithm~\cite{revach2022kalmannet} and its variants
 ~\cite{RTSNet_TSP, buchnik2023latent, choi2023split,chen2025maml},  
 augments the \ac{kg} computation with a \ac{dnn}.

\begin{figure}
    \centering
    \includegraphics[width=\figWidth]{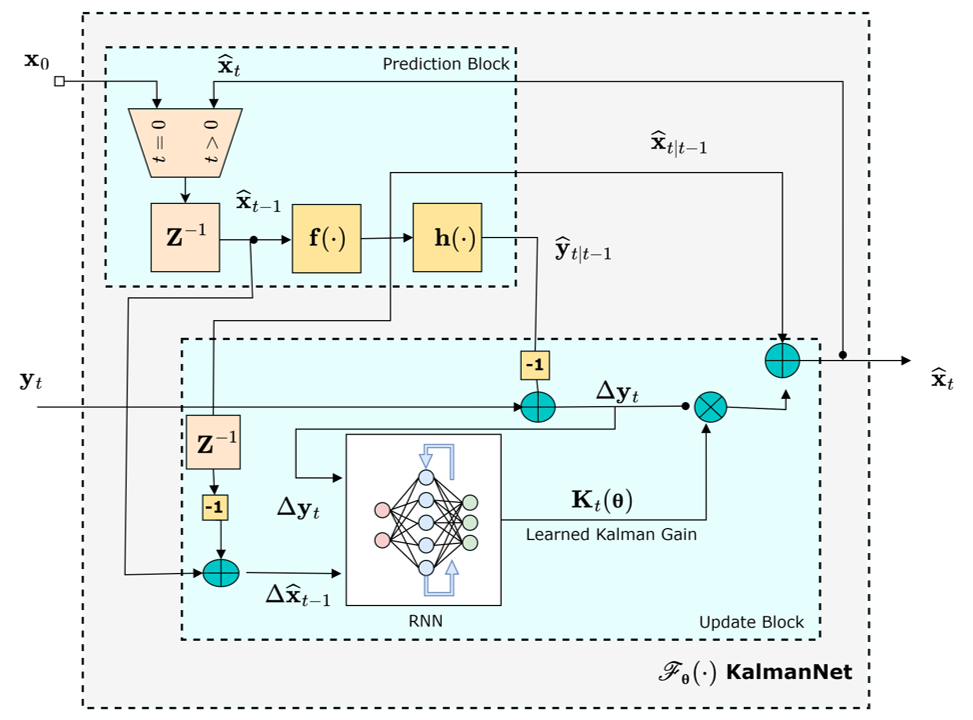}
    \caption{\ac{ekf} with learned \ac{kg} illustration.}
    \label{fig:KalmanNetArch}
\end{figure}

{\bf Architecture:}
Consider a dynamic system represented as a \ac{ss} model in which one has only a (possibly approximated) model of the time-invariant state evolution function $f(\cdot)$  and the observation function $h(\cdot)$. An \ac{ekf} with a learned \ac{kg} (illustrated in Fig.~\ref{fig:KalmanNetArch}) employs a \ac{dnn} with parameters $\dnnParam$ to compute the \ac{kg} for each time instant $t$, denoted $\Kgain_t(\dnnParam)$. Using this learned \ac{kg}, an \ac{ekf} is applied, predicting only the first-order moments via \eqref{eqn:predict_NL}, and estimating the state as 
\begin{equation}
\label{eqn:InferenceKalmanNet}
    \hat{\myVec{x}}_t = \hat{\myVec{x}}_{t|t-1} + \Kgain_t(\dnnParam) (\myVec{y}_t - \hat{\myVec{y}}_{t|t-1}).
\end{equation}  

The  architecture does not explicitly track the second-order moments, that are needed for the \ac{kg}, and thus the learned computation must do so implicitly. Accordingly, the \ac{dnn}  has to 
    $1)$ Process input features that are informative of the noise signals; and 
    $2)$ Possess some internal memory capabilities. 
To meet the first requirement, the input features processed by the \ac{dnn} typically include $(i)$ differences in the observations, e.g., $\Delta \myVec{y}_t = \myVec{y}_t - \hat{\myVec{y}}_{t|t-1}$; 
and $(ii)$ differences in the estimated state, such as $\Delta \hat{\myVec{x}}_{t-1} = \hat{\myVec{x}}_{t-1} - \hat{\myVec{x}}_{t-1|t-2}$. 
To provide internal memory, architectures based on \acp{rnn}~\cite{revach2022kalmannet,choi2023split}, or even transformers~\cite{wang2024nonlinear} can be employed. For instance, a simple \ac{fc}-\ac{rnn}-\ac{fc} architecture was proposed in \cite{revach2022kalmannet}, as well as a more involved interconnection of \acp{rnn}, while \cite{choi2023split} used two \acp{rnn} -- one for tracking $\myMat{\Sigma}_{t|t-1}$ and one for tracking $\myMat{S}_{t|t-1}^{-1}$ -- combined into the \ac{kg} via \eqref{eq:FWGain}. 

While the above architecture is formulated for filtering, it extends to smoothing. Particularly, when smoothing via the \ac{rts} algorithm~\cite{sarkka2023bayesian}
employs a \ac{kf} for a forward pass, followed by an additional backward pass  \eqref{eq:BW_update} that uses the backward gain matrix $\overleftarrow{\Kgain}_t$ given in \eqref{eq:bw_gain}.  Consequently, the \ac{dnn} augmented methodology above extends to smoothing by introducing an additional \ac{kg} \ac{dnn} that learns to compute $\overleftarrow{\Kgain}_t$. Moreover, the single forward-backward pass of the  \ac{rts} algorithm is not necessarily \ac{mse} optimal in non-linear \ac{ss} models, the learned \ac{rts} can be applied in multiple iterations, using the estimates produced at a given forward-backward pass be used as the input to the following pass~\cite{RTSNet_TSP}. Here, the fact that each pass is converted into a \ac{ml} architecture allows to learn a different forward and backward gain \acp{dnn} for each pass, as a form of deep unfolding~\cite{shlezinger2022model}
 
{\bf Training:} 
KalmanNet and its variants use \acp{dnn} to compute the \ac{kg}. While there is no `ground-truth` \ac{kg} when deviating from linear Gaussian \ac{ss} models, the overall augmented state estimation algorithm is trainable in a supervised manner. 
\color{NewColor}
Particularly, given a labeled data set as in \eqref{eqn:DataSet}, a candidate loss measure is 
\begin{equation}
    \label{eqn:lossJoint2}
    \mathcal{L}_{\mySet{D}_{\rm s}}(\dnnParam) = \frac{1}{\Ntraining T}\sum_{i=1}^{\Ntraining}\sum_{t=1}^{T} \left\|\hat{\myVec{x}}_t\big(\myVec{y}_{1:t}^{(i)}; \dnnParam\big) -  \myVec{x}_t^{(i)}\right\|_2,
\end{equation}
where  $\hat{\myVec{x}}_t\big(\myVec{y}_{1:t}; \dnnParam\big)$ is the state estimated using the augmented \ac{ekf} with parameters $\dnnParam$ and observations $\myVec{y}_{1:t}$. 
\color{black}

While KalmanNet is designed to be trained from labeled data via \eqref{eqn:lossJoint2}, in some settings it can also be trained without providing it with ground-truth state labels. This can be achieved via the following approaches:
\begin{itemize}
    \item {\em Observation Prediction:} The fact that the \ac{dnn}-augmented algorithm preserves the interpretable operation of Kalman-type tracking can be leveraged for unsupervised learning, i.e., learning from  data  of the form
    \begin{equation}
    \label{eqn:UnsupData}
        \mySet{D}_{\rm u}    = \big\{\myVec{y}_{1:T}^{(i)}\big\}_{i=1}^{\Ntraining}.
    \end{equation}
    Specifically, as Kalman-type algorithms with time-invariant \ac{ss} models internally predict the next observation as $\hat{\myVec{y}}_{t+1|t} = h \big( f ( \hat{\myVec{x}}_t) \big)$ in \eqref{eqn:predict_obs}, a possible unsupervised loss is~\cite{revach2021unsupervised}
    \begin{equation}
    \label{eqn:lossJoint3}
    \mathcal{L}_{\mySet{D}_{\rm u}  }(\dnnParam) = \frac{1}{\Ntraining T}\sum_{i=1}^{\Ntraining}\sum_{t=0}^{T-1} \left\|h \big( f \big(\hat{\myVec{x}}_t\big(\myVec{y}_{1:t}^{(i)}; \dnnParam\big)\big)\big) -  \myVec{y}_{t+1}^{(i)}\right\|_2.
    \end{equation}
    \item {\em Downstream Task:} 
    Often, in practice, state estimation is carried out as part of an overall processing chain, with some downstream tasks. In various applications, one can evaluate an estimated state without requiring a ground truth value. When such evaluation is written (or approximated) as a function that is differentiable with respect to the estimated state, it can be used as a training loss. This approach was shown to enable unsupervised learning of KalmanNet when integrated in stochastic control systems, where it is combined with a linear quadratic regulator~\cite{putri2023data}, as well as in financial pairs trading, where the states tracked are financial features used for trading policies~\cite{milstein2024neural}.
\end{itemize}

\begin{tcolorbox}[float*=t,
    width=\boxWidth,
	toprule = 0mm,
	bottomrule = 0mm,
	leftrule = 0mm,
	rightrule = 0mm,
	arc = 0mm,
	colframe = myblue,
	colback = mypurple,
	fonttitle = \sffamily\bfseries\large,
	title = Uncertainty Extraction from Learned \ac{kg}]	
	\label{Box:KalmanNet}
\begin{wrapfigure}{r}{7cm} 
    \centering
    \vspace{-0.4cm}
    \includegraphics[width=7cm]{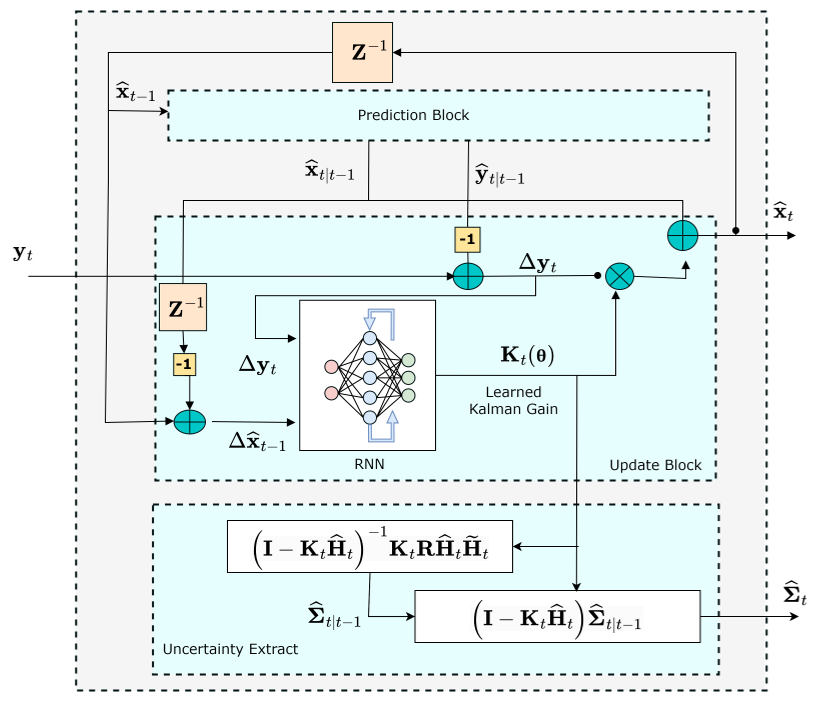} 
    \caption{\ac{ekf} with learned \ac{kg} and covariance extraction illustration.} 
    \label{fig:CovExtract}
\end{wrapfigure}
 Kalman-type algorithms use both the prior covariance $\myMat{\Sigma}_{t|t-1}$ as well as the \ac{kg} to compute the posterior error covariance $\myMat{\Sigma}_t$, which represents the estimation uncertainty. Particularly, by \eqref{eqn:update2} and \eqref{eq:FWGain}, the \ac{ekf} computes the error covariance as  
 \begin{subequations}
     \label{eqn:ErrorCovEst}
\begin{align} 
    \myMat{\Sigma}_{t} &= (\myMat{I}-  \myMat{K}_{t} \hat{\myMat{H}}_{t}) \myMat{\Sigma}_{t|t-1}.  \label{eq:kf_cov_upd} 
\end{align} 
Consequently, despite the fact that \ac{ai}-aided filters with learned \ac{kg} do not explicitly track the second-order moments, in some cases, these can still be recovered via  \eqref{eq:kf_cov_upd}, as shown in \cite{Dahan2024uncertainty}.

In particular, the architecture provides the learned \ac{kg}  as $\myMat{K}_{t}(\dnnParam)$. Accordingly, when   the matrix $\Tilde{\myMat{H}}_{t} = \big(\hat{\myMat{H}}_{t}^\top\hat{\myMat{H}}_{t}\big)^{-1}$ exists, then,  combining \eqref{eqn:obs_2} and \eqref{eq:FWGain},  the prior covariance can be recovered via 
\begin{equation}
\label{eqn:PriorErrorCov}
       {\myMat{\Sigma}}_{t|t-1}= (\myMat{I} - {\myMat{K}}_t(\dnnParam)  \hat{\myMat{H}}_{t})^{-1}{\myMat{K}}_t(\dnnParam)  \myMat{R}\hat{\myMat{H}}_{t}\Tilde{\myMat{H}}_{t}.
\end{equation}
 \end{subequations}
 
This indicates that when one accurately computes $\hat{\myMat{H}}_{t}$ (via \eqref{eqn:update_NL}) and has knowledge of the measurement noise covariance $\myMat{R}$, then the learned \ac{kg} can be used to recover the error covariance via \eqref{eq:kf_cov_upd} and \eqref{eqn:PriorErrorCov}, as illustrated in Fig.~\ref{fig:CovExtract}. The extracted covariance can be encouraged to match the empirical estimation error in training by, e.g., adding an additional loss term that penalizes uncertainty extraction, or by imposing a Gaussian prior on the error and minimizing its likelihood, see \cite{Dahan2024uncertainty}. 

\end{tcolorbox}

{\bf Discussion:}
The design of \ac{dnn}-aided Kalman-type algorithms by augmenting the \ac{kg} computation is particularly suitable for tackling identified challenges in \ref{itm:FuncAccuracy}-\ref{itm:InferenceSpeed}. Unlike purely end-to-end \acp{dnn} designed for generic processing of time-sequences, augmenting the \ac{kg} allows leveraging domain knowledge of the state evolution and observation models, as these are utilized in the prediction step. However, as the following processing of these predictions utilizes a \ac{dnn} that is trained based on the accuracy of the overall algorithm, mismatches, and approximation errors are learned to be corrected, thus fully tackling~\ref{itm:FuncAccuracy}. As the \ac{dnn} augmentation bypasses the need to track second-order moments, the resulting algorithm does not require knowledge of the distribution of the noises. In fact, the resulting filter is not linear (as the \ac{kg} depends on the observations), allowing to learn non-linear state estimators that are suitable for non-Gaussian \ac{ss} models (\ref{itm:Stochasticity}) and non-linear dynamics (\ref{itm:NonLinear}). Finally, as the computation of the \ac{kg} and the propagation of the second-order moments induces most of the latency in Kalman-type algorithms for non-linear \ac{ss} models, replacing these computations with a compact \ac{dnn} often leads to more rapid inference (\ref{itm:InferenceSpeed}) \cite{RTSNet_TSP}.

The fact that the resulting state estimator is converted into a trainable discriminative \ac{ml} architecture makes it natural to be combined with \ac{dnn}-based pre-processing (as in Fig.~\ref{fig:SeperateDNN}), e.g., for processing images or high dimensional data. Particularly, the learned state estimator can be trained jointly with the pre-processing stage, thus having it learn latent features that are most useful for processing with the learned filter, without requiring one to approximate the distribution of the latent measurement noise~\cite{buchnik2023latent}.

Unlike state estimation based on end-to-end \acp{dnn}, which also tackles \ref{itm:FuncAccuracy}-\ref{itm:InferenceSpeed} (as discussed in the previous section), the usage of Kalman-type algorithms as an inductive bias allows to also preserve some of the desirable properties of model-based state estimators. In particular, the interpretable operation is preserved, in the sense that the internal features are associated with concrete meaning (\ref{itm:interpretability}), which can be exploited for, e.g., unsupervised learning. Moreover, while the error covariance is not explicitly tracked, in some cases it can actually be recovered from the learned \ac{kg} (see box entitled {\em `Uncertainty Extraction from Learned \ac{kg}`} on Page~\pageref{Box:KalmanNet}), thus providing \ref{itm:uncertainty} to some extent. 
In addition, the fact that only an internal computation is learned allows utilizing relatively compact \acp{dnn}, striking a balance between the excessive complexity of end-to-end \acp{dnn} and the relatively low one of model-based methods (\ref{itm:complexity}).

The core limitations of designing \ac{ai}-aided state estimator by augmenting the \ac{kg} computations are associated with its adaptivity. The design is particularly geared towards time-invariant state evolution and observation functions. As the \ac{kg} computation is coupled with these models, any deviation, even known ones, is expected to necessitate re-training. This can be alleviated under some forms of variations using hypernetworks~\cite{ni2023adaptive}, i.e., having an additional \ac{dnn} that updates the \ac{kg} \ac{dnn}, while inducing some complexity increase. Moreover, the architecture is designed for supervised learning. The ability to train in an unsupervised manner via \eqref{eqn:lossJoint3} relies on full knowledge of $f(\cdot)$ and $h(\cdot)$, thus not being applicable in the same range of settings as its supervised counterpart. The same also holds for uncertainty extraction via \eqref{eqn:ErrorCovEst}, which requires some additional domain knowledge compared to that needed for merely tracking the state via \eqref{eqn:InferenceKalmanNet}. 




The introduced learned \ac{kg} algorithms calculate the complete gain by a \ac{dnn}. However, if a Kalman-type filter, such as the unscented Kalman filter or divided-difference filter, 
depends on a \textit{user-defined} scaling parameter, a \ac{dnn} can be used for the parameter prediction (instead of the complete gain calculation) 
\cite{FaShBaPhBlCh:24}. As a consequence, such \ac{dnn}-reasoned scaling parameter affects all elements of the \ac{kg} and is able, up to a certain extent, to compensate for model discrepancies or linearisation errors. This \ac{dnn}-augmented filter inherently provides the estimate error covariance matrix, but under the assumption of the known \ac{ss} model \eqref{eq:NLssModel}.

\begin{tcolorbox}[float*=t,
    width=\boxWidth,
	toprule = 0mm,
	bottomrule = 0mm,
	leftrule = 0mm,
	rightrule = 0mm,
	arc = 0mm,
	colframe = myblue,
	colback = mypurple,
	fonttitle = \sffamily\bfseries\large,
	title = DANSE Architecture]	
	\label{Box:DANSE}
\begin{wrapfigure}{r}{10cm} 
    \centering
    \vspace{-0.4cm}
    \includegraphics[width=10cm]{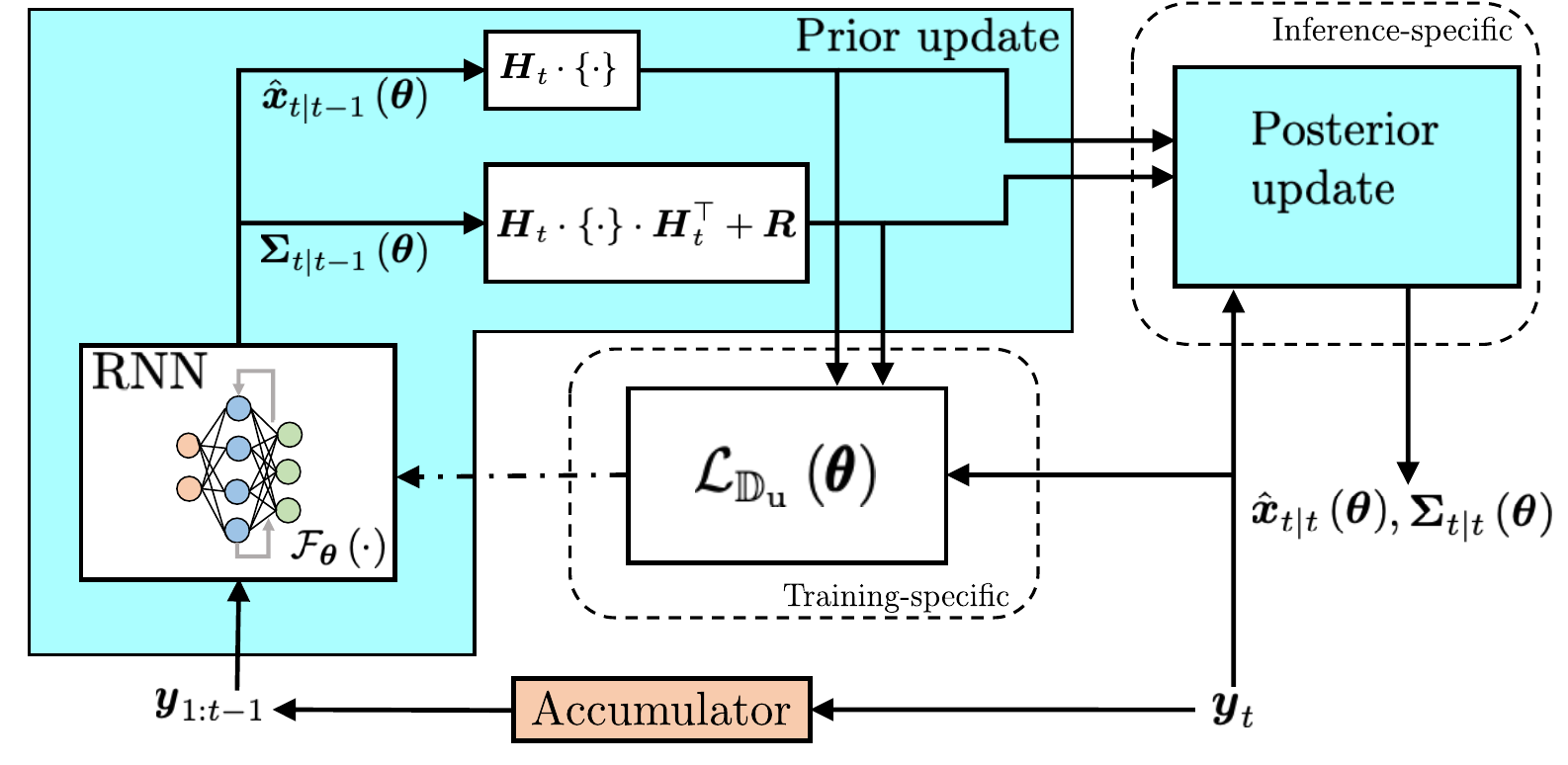} 
    \caption{Diagram of  DANSE  at time $t$, highlighting both training-specific and inference-specific blocks \cite{ghosh2023danse}. The dash-dotted line represents the gradient flow to $\dnnFunc$ during training, solid lines indicate information flow during training/inference. } 
    \label{fig:DANSEfig}
\end{wrapfigure}
The DANSE method utilizes an RNN for recursively processing the input sequence $\myVec{y}_{1:t-1}$ \cite{ghosh2023danse}. An example of this sequential processing at time $t$ is shown in Fig. \ref{fig:DANSEfig}. One can use prominent RNNs for sequential processing, such as GRUs or LSTMs. 
 In DANSE, GRU was used due to simplicity and good expressive power compared to vanilla RNNs. Specifically, for modeling the mean vector $\hat{\myVec{x}}_{t \vert t-1}\left(\dnnParam\right)$ and diagonal covariance matrix ${\myMat{\Sigma}}_{t \vert t-1}\left(\dnnParam\right)$ of the parameterized Gaussian prior in \eqref{eq:DANSERNNprior}, the hidden state of the GRU was nonlinearly transformed using feed-forward networks with suitable activation functions. In Fig. \ref{fig:DANSEfig}, the `Accumulator' block is shown for ease of understanding; in practice, the block is embedded in the recursive operation of the RNN. 
\end{tcolorbox}

\subsubsection{Learned State Estimation}
\label{ssec:discriminative_no_ss}
Another class of integrated DNN architectures includes methods that learn the task of state estimation in a data-driven manner using the known observation model without any knowledge of the state evolution model. The state evolution model in \eqref{eq:ssModelState} may not be linear or require $\myVec{v}_t$ to be Gaussian noise. A candidate approach in this category is the data-driven nonlinear state estimation (DANSE) method that provides a closed-form posterior of the underlying state using linear state measurements under Gaussian noise \cite{ghosh2023danse}. The noisy measurements follow  the linear observation model similar to \eqref{eq:LinssModelObs}, namely,
\begin{equation}\label{eq:measurementsys_danse}
    \myVec{y}_t = \myMat{H}_{t} \myVec{x}_t + \myVec{w}_t,
\end{equation}
where the measurement noise is  $\myVec{w}_t \sim \mathcal{N}\left(\myVec{0}, \myMat{R}\right)$ with known covariance $\myMat{R}$. The matrix $\myMat{H}_{t}$ is assumed to be full column rank and known $\forall t$. Note that, $p\left(\myVec{y}_t \vert \myVec{x}_t \right) = \mathcal{N}\left(\myMat{H}_t \myVec{x}_t, \myMat{R}\right)$.  

{\bf Architecture:} The core of the DANSE method relies on the sequential modeling capability of \acp{rnn} \cite[Ch. 10]{goodfellow2016deep}. DANSE consists of an RNN with parameters $\dnnParam$ that parameterizes a Gaussian prior on $\myVec{x}_t$ given $\myVec{y}_{1:t-1}$ at a given time $t$. Concretely, the prior distribution $p\left(\myVec{x}_t \vert \myVec{y}_{1:t-1}; \dnnParam \right)$ is 
\ifsingle
\begin{align}\label{eq:DANSERNNprior}
    &p\left(\myVec{x}_t \vert \myVec{y}_{1:t-1}; \dnnParam \right) = \mathcal{N}\left(\hat{\myVec{x}}_{t \vert t-1}\left(\dnnParam\right), \myMat{\Sigma}_{t \vert t-1}\left(\dnnParam\right)\right) 
    \quad \text{s.t. }\lbrace \hat{\myVec{x}}_{t \vert t-1}\left(\dnnParam\right), \myMat{\Sigma}_{t \vert t-1}\left(\dnnParam\right) \rbrace = \dnnFunc(\myVec{y}_{t-1}),
\end{align}   
\else
\begin{align}\label{eq:DANSERNNprior}
    &p\left(\myVec{x}_t \vert \myVec{y}_{1:t-1}; \dnnParam \right) = \mathcal{N}\left(\hat{\myVec{x}}_{t \vert t-1}\left(\dnnParam\right), \myMat{\Sigma}_{t \vert t-1}\left(\dnnParam\right)\right) \notag \\
    &\text{s.t. }\lbrace \hat{\myVec{x}}_{t \vert t-1}\left(\dnnParam\right), \myMat{\Sigma}_{t \vert t-1}\left(\dnnParam\right) \rbrace = \dnnFunc(\myVec{y}_{t-1}),
\end{align}   
\fi
where $\dnnFunc(\myVec{y}_{t-1})$ refers to the RNN that recursively processes the sequence of past observations $\myVec{y}_{1:t-1}$ as described in \eqref{eqn:RNNb}, \eqref{eqn:RNNa}. Also, $\hat{\myVec{x}}_{t \vert t-1}\left(\dnnParam\right) \text{ and } \myMat{\Sigma}_{t \vert t-1}\left(\dnnParam\right)$ denote the mean vector and the covariance matrix respectively of the RNN-based Gaussian prior at time $t$. The covariance matrix $\myMat{\Sigma}_{t \vert t-1}\left(\dnnParam\right)$ can be designed to be full or diagonal. More accurately, the hidden state of the RNN in \eqref{eq:DANSERNNprior} at time $t$ is non-linearly transformed using feed-forward networks with appropriate activation functions to obtain $\hat{\myVec{x}}_{t \vert t-1}\left(\dnnParam\right) \text{ and } \myMat{\Sigma}_{t \vert t-1}\left(\dnnParam\right)$ \cite{ghosh2023danse}.
In \cite{ghosh2023danse}, DANSE uses GRU for implementation purposes owing to its simplicity and modeling capability. 

Using the fact that the measurement system in \eqref{eq:measurementsys_danse} is linear and also Gaussian, we have $p\left(\myVec{y}_t \vert \myVec{x}_t\right) = \mathcal{N}\left(\myMat{H}_t \myVec{x}_t, \myMat{R}\right)$. This allows the use of the representation in equation \eqref{eqn:CK1a} to obtain $p\left(\myVec{x}_t \vert \myVec{y}_{1:t}; \dnnParam\right)$ in closed-form \cite{ghosh2023danse}. One can show that the posterior distribution 
\ifsingle
holds $p\left(\myVec{x}_t \vert \myVec{y}_{1:t}; \dnnParam\right)=\mathcal{N}( \hat{\myVec{x}}_{t \vert t}(\dnnParam), {\myMat{\Sigma}}_{t \vert t}(\dnnParam))$  with 
\begin{align}\label{eq:DANSERNNPosterior}
    &\hat{\myVec{x}}_{t \vert t}(\dnnParam) 
    = \hat{\myVec{x}}_{t \vert t-1}(\dnnParam) + \myMat{K}_{t}^{\prime}\left(\dnnParam\right) \Delta\myVec{y}_t \left(\dnnParam\right), 
    \quad 
    {\myMat{\Sigma}}_{t \vert t}(\dnnParam) 
     = {\myMat{\Sigma}}_{t \vert t-1}(\dnnParam) 
    -  \myMat{K}_{t}^{\prime}\left(\dnnParam\right)\myMat{S}_{t \vert t-1}\left(\dnnParam\right) \left(\myMat{K}_{t}^{\prime}\left(\dnnParam\right)\right)^{\top}.
\end{align}
\else
$p\left(\myVec{x}_t \vert \myVec{y}_{1:t}; \dnnParam\right)$ is also Gaussian with 
\begin{align}\label{eq:DANSERNNPosterior}
    p(\myVec{x}_t|\myVec{y}_{1:t}; \dnnParam) &=  \mathcal{N}( \hat{\myVec{x}}_{t \vert t}(\dnnParam), {\myMat{\Sigma}}_{t \vert t}(\dnnParam)), \notag \\
    \hat{\myVec{x}}_{t \vert t}(\dnnParam) 
    &= \hat{\myVec{x}}_{t \vert t-1}(\dnnParam) + \myMat{K}_{t}^{\prime}\left(\dnnParam\right) \Delta\myVec{y}_t \left(\dnnParam\right), \notag\\
    {\myMat{\Sigma}}_{t \vert t}(\dnnParam) 
     &= {\myMat{\Sigma}}_{t \vert t-1}(\dnnParam) 
    -  \myMat{K}_{t}^{\prime}\left(\dnnParam\right)\myMat{S}_{t \vert t-1}\left(\dnnParam\right) \left(\myMat{K}_{t}^{\prime}\left(\dnnParam\right)\right)^{\top}.
\end{align}
\fi
Notably, \eqref{eq:DANSERNNPosterior} resembles the Kalman update equations in \eqref{eqn:update}, \eqref{eq:FWGain} where 
\begin{align}\label{eq:DANSERNNPosteriorDetails}
 \myMat{K}_{t}^{\prime}\left(\dnnParam\right) &\triangleq {\myMat{\Sigma}}_{t \vert t-1}(\dnnParam)\myMat{H}_t^{\top}\myMat{S}^{-1}_{t \vert t-1}\left(\dnnParam\right), \notag \\
    \myMat{S}_{t \vert t-1}\left(\dnnParam\right) &\triangleq \myMat{H}_t\myMat{\Sigma}_{t \vert t-1}\left(\dnnParam\right)\myMat{H}_t^{\top} + \myMat{R}, \notag \\
    \Delta\myVec{y}_t\left(\dnnParam\right)  &\triangleq \myVec{y}_t - \myMat{H}_t\hat{\myVec{x}}_{t \vert t-1}(\dnnParam). 
\end{align}
Note that $\myMat{K}_{t}^{\prime}\left(\dnnParam\right)$ is conceptually similar as the traditional Kalman gain term in \eqref{eq:FWGain}, but computed in a data-driven manner, and that the computation approach is  different from the one illustrated in Fig. \ref{fig:KalmanNetArch} due to \eqref{eq:DANSERNNprior}. 

\begin{figure}[t]
    \centering
    \includegraphics[width=\figWidth]{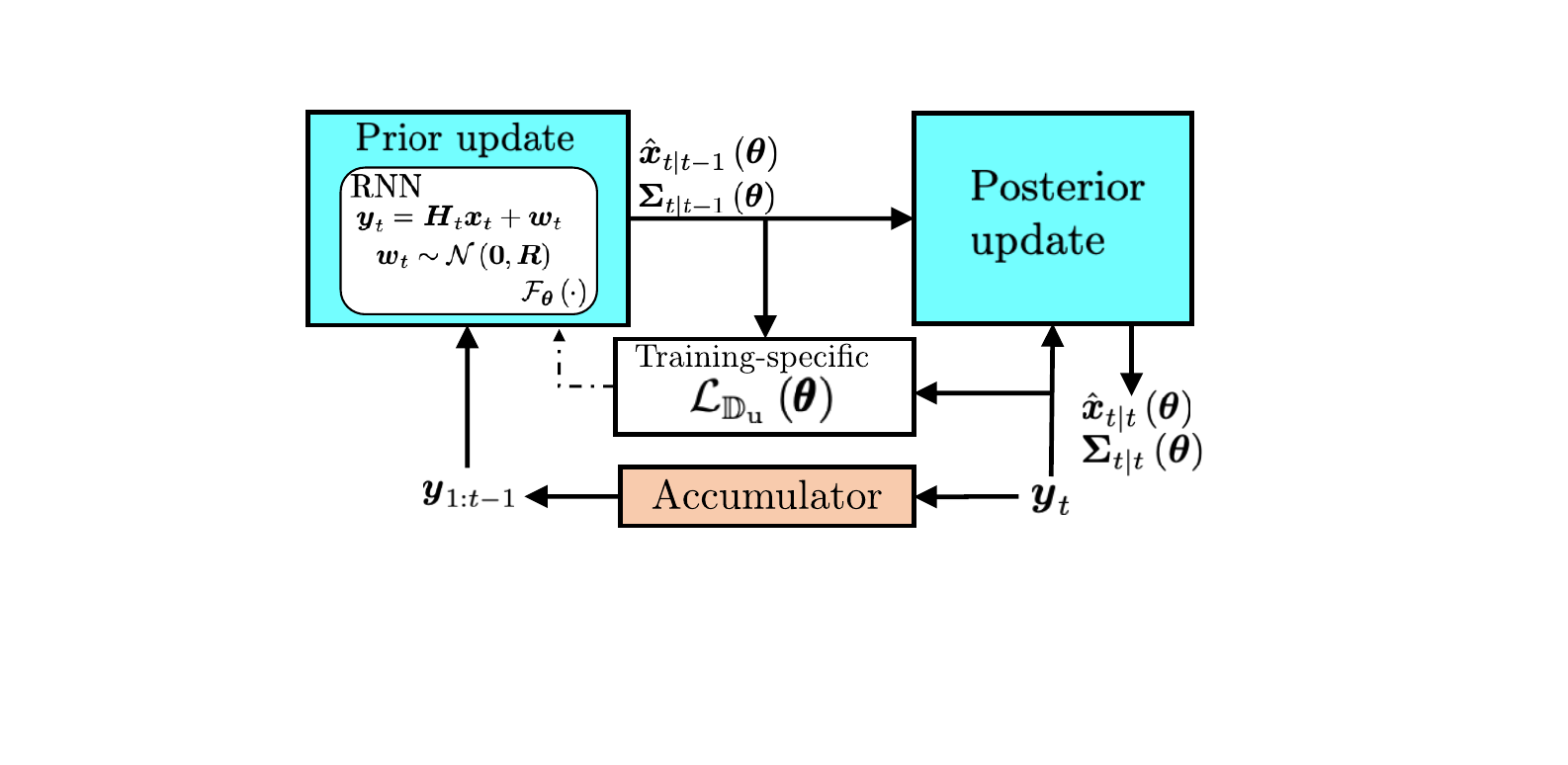}
    \caption{Simplified schematic of DANSE \cite{ghosh2023danse}.
    }
    \label{fig:danse_architecture}
\end{figure}

In the backdrop of KF, a crucial aspect of DANSE is that there is no Gaussian propagation of the posterior in \eqref{eq:DANSERNNPosterior} to the prior because DANSE does not use any state evolution model like \eqref{eq:ssModelState}.  The parameters $\hat{\myVec{x}}_{t \vert t-1} \left(\dnnParam\right), \myMat{\Sigma}_{t \vert t-1} \left(\dnnParam\right)$ of the prior are obtained directly from the RNN, which does not use any explicit state evolution model as in \eqref{eq:ssModelState}, or require any knowledge regarding the same, e.g., first-order Markovian, Gaussian process noise, etc. A simplified schematic of DANSE is shown in Fig. \ref{fig:danse_architecture} with a detailed schematic and details regarding architectural choices present in the box `DANSE Architecture' on Page \pageref{Box:DANSE}. 


{\bf Training:} The parameters $\dnnParam$ in DANSE are learned in an unsupervised manner using a training dataset consisting of only noisy measurement trajectories $\mySet{D}_{\rm u}   = \lbrace \myVec{y}_{1:T}^{(i)} \rbrace_{i=1}^{N}$, where $N$ is the number of training samples, where every $i$'th sample is assumed to have the same trajectory length $T$. The learning mechanism is based on maximizing the likelihood of the dataset $\mySet{D}$, where one computes the joint likelihood of a full sequence $\myVec{y}_{1:T}$. 
In order to achieve this, one first calculates the conditional marginal distribution $p\left(\myVec{y}_{t} \vert \myVec{y}_{1:t-1}; \dnnParam\right)$ using the Chapman-Kolmogorov equation \eqref{eqn:CK1b} as follows
\ifsingle
\begin{align}\label{eq:DANSERNNMarginal}
p(\myVec{y}_t | \myVec{y}_{1:t-1}; \dnnParam) 
&= \displaystyle\int p(\myVec{y}_t | \myVec{x}_t )p(\myVec{x}_t |  \myVec{y}_{1:t-1}; \dnnParam ) \, d\myVec{x}_t 
= \mathcal{N}(\myMat{H}_t \hat{\myVec{x}}_{t \vert t-1}(\dnnParam),  \myMat{H}_t {\myMat{\Sigma}}_{t \vert t-1}(\dnnParam) \myMat{H}_t^{\top}+\myMat{R}) \notag \\
&= \mathcal{N}(\myMat{H}_t \hat{\myVec{x}}_{t \vert t-1}(\dnnParam),  \myMat{S}_{t \vert t-1}\left(\dnnParam\right)),
\end{align}
\else
\begin{align}\label{eq:DANSERNNMarginal}
p(\myVec{y}_t | \myVec{y}_{1:t-1}; \dnnParam) 
&= \displaystyle\int p(\myVec{y}_t | \myVec{x}_t )p(\myVec{x}_t |  \myVec{y}_{1:t-1}; \dnnParam ) \, d\myVec{x}_t \notag \\ 
&= \mathcal{N}(\myMat{H}_t \hat{\myVec{x}}_{t \vert t-1}(\dnnParam),  \myMat{H}_t {\myMat{\Sigma}}_{t \vert t-1}(\dnnParam) \myMat{H}_t^{\top}+\myMat{R}) \notag \\
&= \mathcal{N}(\myMat{H}_t \hat{\myVec{x}}_{t \vert t-1}(\dnnParam),  \myMat{S}_{t \vert t-1}\left(\dnnParam\right)),
\end{align}
\fi
where we  $\myMat{S}_{t \vert t-1}\left(\dnnParam\right)$ is defined in \eqref{eq:DANSERNNPosteriorDetails}. Then for the sequence $\myVec{y}_{1:T}$, one can calculate the joint likelihood as 
\begin{equation}\label{eq:DANSEjointlikelihood}
    p\left(\myVec{y}_{1:T}; \dnnParam\right) = \prod_{t=1}^{T} p(\myVec{y}_t | \myVec{y}_{1:t-1}; \dnnParam). 
\end{equation}
Thus, using $\mySet{D}_{\rm u}  $, \eqref{eq:DANSERNNMarginal} and \eqref{eq:DANSEjointlikelihood}, the optimization problem can be formulated as maximization of the logarithm of the joint log-likelihood of $\mySet{D}_{\rm u}$ as follows   
\begin{align}\label{eq:DANSEOptimizationProblem}
    \max_{\dnnParam} \log \prod_{i=1}^{N} \prod_{t=1}^{T} p\left( \myVec{y}_t^{(i)} | \myVec{y}_{1:t-1}^{(i)} ; {\dnnParam} \right) 
    &= \min_{\dnnParam} \mathcal{L}_{\mySet{D}_{\rm u}  }\left({\dnnParam}\right),
\end{align}
where the loss term $\mathcal{L}_{\mySet{D}_{\rm u}  }\left({\dnnParam}\right)$ is obtained using \eqref{eq:DANSERNNMarginal} as follows
\ifsingle
\begin{align}\label{eq:DANSELossfm}
\mathcal{L}_{\mySet{D}_{\rm u}  }\left( {\dnnParam}\right)  
&= \sum_{i=1}^{N} \sum_{t=1}^{T} \Bigg\lbrace \frac{n}{2}\log 2 \pi  +  \frac{1}{2}\log \text{det} \left(\myMat{S}^{(i)}_{t \vert t-1}\left(\dnnParam\right) \right)  + \frac{1}{2} \Vert \myVec{y}_t - \myMat{H}_t \hat{\myVec{x}}^{(i)}_{t \vert t-1}(\dnnParam)\Vert_{\left(\myMat{S}^{(i)}_{t \vert t-1}\left(\dnnParam\right)\right)^{-1}}^2 \Bigg\rbrace.
\end{align}
\else
\begin{align}\label{eq:DANSELossfm}
\mathcal{L}_{\mySet{D}_{\rm u}  }\left( {\dnnParam}\right)  
&= \sum_{i=1}^{N} \sum_{t=1}^{T} \Bigg\lbrace \frac{n}{2}\log 2 \pi  +  \frac{1}{2}\log \text{det} \left(\myMat{S}^{(i)}_{t \vert t-1}\left(\dnnParam\right) \right) \notag \\
&+ \frac{1}{2} \Vert \myVec{y}_t - \myMat{H}_t \hat{\myVec{x}}^{(i)}_{t \vert t-1}(\dnnParam)\Vert_{\left(\myMat{S}^{(i)}_{t \vert t-1}\left(\dnnParam\right)\right)^{-1}}^2 \Bigg\rbrace.
\end{align}
\fi
In \eqref{eq:DANSELossfm}, $\lbrace \hat{\myVec{x}}^{(i)}_{t \vert t-1}(\dnnParam), \myMat{\Sigma}^{(i)}_{t \vert t-1}\left(\dnnParam\right)\rbrace $ are the Gaussian prior parameters  \eqref{eq:DANSERNNprior} obtained using the $i$'th sample $\myVec{y}_{1:t-1}^{(i)}$ as input at time $t$.
The parameters $\dnnParam$ are thus learned by minimizing $\mathcal{L}_{\mySet{D}_{\rm u}  }\left({\dnnParam}\right)$ with respect to $\dnnParam$ in \eqref{eq:DANSEOptimizationProblem}. Practically, this is achieved by using mini-batch stochastic gradient descent \cite[Sec. II-C]{ghosh2023danse}. 

{\bf Discussion:} The DANSE approach can  address the challenges associated with model-based KF-type algorithms explained in \ref{itm:FuncAccuracy}-\ref{itm:InferenceSpeed}. DANSE partly addresses  \ref{itm:FuncAccuracy} as it does not require knowledge of the state evolution model \eqref{eq:ssModelState} or the involved process noise in \eqref{eq:ssModelState}. However, it requires knowledge of the linear observation model shown in \eqref{eq:measurementsys_danse}, where  $\myMat{H}_t$ should be full-rank and noise covariance $\myMat{R}$ should be known \textit{a priori}. The parameterization of the Gaussian prior in \eqref{eq:DANSERNNprior} ensures that DANSE can capture long-term temporal dependency in the state model so that one is not constrained to state estimation for Markovian SS models. This is also mentioned \cite{ghosh2023danse} where  DANSE is introduced in a `Model-free process' setting. In \cite{ghosh2023danse}, the authors define a `Model-free process' as a process where the knowledge of the state evolution model is absent. This also partly tackles \ref{itm:Stochasticity} since there is no requirement on the distribution of $\myVec{v}_t$ in \eqref{eq:ssModelState} to be Gaussian. At the same time, it is required that $\myVec{w}_t$ is Gaussian with a fixed covariance $\myMat{R}$ as shown in \eqref{eq:measurementsys_danse}. 

The use of \acp{rnn} and the parameterization of the Gaussian prior also ensures that one can learn a nonlinear state estimation method for nonlinear SS models similar to KalmanNet. This in turn helps to tackle the challenge described in \ref{itm:NonLinear}. The training of DANSE, as explained in the previous paragraph, is unsupervised and offline as it requires access to a dataset $\mySet{D}_{\rm u}  $ consisting of noisy measurements. Once the training is completed, the inference step in DANSE is causal and separated from the offline training step. This ensures that one has a rapid inference at test time. This helps address \ref{itm:InferenceSpeed}, as DANSE does not require any linearization of the state evolution model as in EKF or sampling sigma points or particles in UKF or PF, respectively. 

 DANSE  is also partly interpretable as its posterior update  in \eqref{eq:DANSERNNPosterior} is tractable and bears similarity to that of the KF  \eqref{eqn:update}. This is ensured by using the linear observation model in \eqref{eq:measurementsys_danse} and the choice of the Gaussian prior for $\myVec{x}_t$ in \eqref{eq:DANSERNNprior}. This ensures tractable posterior updates, resonating with the feature \ref{itm:interpretability} regarding the interpretability of conventional model-based approaches using first and second-order moments. Furthermore, the advantage \ref{itm:uncertainty} regarding providing uncertainty estimates is also present in DANSE as we have both prior and posterior covariance estimates in \eqref{eq:DANSERNNprior}, \eqref{eq:DANSERNNPosterior}. As mentioned earlier, the key difference compared to model-based approaches is the absence of propagating the posterior moments to the prior at the next time point. 

Lastly, it is worth noting that DANSE cannot immediately adapt to underlying changes in the SS model and would require re-training. This also applies to changes in the observation model as it requires complete knowledge of the same. Hence DANSE does not have the advantage \ref{itm:Adaptability}, which is inherent in model-based approaches and AI-aided approaches that utilize the knowledge of the state evolution model or additional hypernetworks. 

\section{\ac{ai}-Augmented \acp{kf} via \ac{ss}-Oriented Learning}
\label{sec:generative}

The second family of \ac{ai}-aided \ac{kf}s uses data to learn, refine, or augment the underlying statistical model using deep learning tools. Then, the data-augmented \ac{ss} can be directly used in Kalman-type state estimation. Instead of learning the state estimation or the specific parameters in the Kalman filter, e.g., KG, the key rationale of \ac{ss}-oriented learning is to exploit the data to obtain a more accurate model. 
At the same time, \ac{ss}-oriented learning preserves explainability of the state and brings associated benefits of the statistical estimation\footnote{Such augmented model can directly be used also for designing a fault-detection or control algorithm.} such as the inherent calculation of the covariance matrix of the estimate error \cite{imbiriba2023augmented}.


{\bf Architecture:}
{\color{NewColor}
The \ac{ss}-oriented \ac{ai}-augmented \acp{kf} generally focus on data-augmented modeling of the state and measurement equations \cite{masti2021learning,gorji2008identification}.
Some  approaches \cite{suykens1995nonlinear, fraccaro2016sequential,arnold2021state,
schoukens2021improved,forgione2020model}
assume the measurement  model known, for which the motivation is twofold: \textit{(i)} the sensors can often be well modeled based on first principles but a state dynamics model is typically approximate and  depends on a user decision (for example, object kinematic can be modeled by nearly constant velocity/acceleration model or a Singer model) \cite{imbiriba2023augmented}; \textit{(ii)} the need to estimate state carrying physical meaning, i.e., to preserve interpretability of state estimate. Interpretability may be lost when both equations are modeled with data augmentation. For convenience and for  compact exposition, the following text will focus on data-augmented modeling of the state dynamics. The concept of data-augmented modeling thus resides in the definition of four models of state dynamics:}
\begin{itemize}
    \item \textit{\Ac{tm}}, or a data generator, is a complex and context-dependent model that cannot be expressed in a finite-dimensional form. The \ac{tm} is thus more of a theoretical construct.
    \item \textit{\Ac{pbm}} is defined by \eqref{eq:NLssModelState} and can be understood as the ``best achievable'' model of the given complexity found from the first principles. 
    \item \textit{\Ac{ddm}} is solely identified from data. The use of \ac{ml} in system identification has been studied for several decades 
    \cite{ljung2020} and the \ac{ddm} can be written as 
    \begin{align}
        \myVec{x}_{t+1} &= g(\myVec{x}_{t};\myVec{\theta}^\dnn) + \myVec{v}_{t}^\ddm, \label{eq:ddModelState}
    \end{align}
    where $\myVec{v}_t^\ddm$ is a process noise, $g(\cdot)$ is a vector-valued function, possibly a \ac{dnn}, which is parametrized by the vector $\myVec{\theta}^\dnn$ designed to minimize the discrepancy between outputs of \eqref{eq:ddModelState} and the \ac{tm}.
    \item The current trend is to consider models that are not purely trained from data, as in \Acp{ddm}, making use of the available information that \Acp{pbm} offer. These hybrid models are discussed in more detail in the sequel.  
\end{itemize}
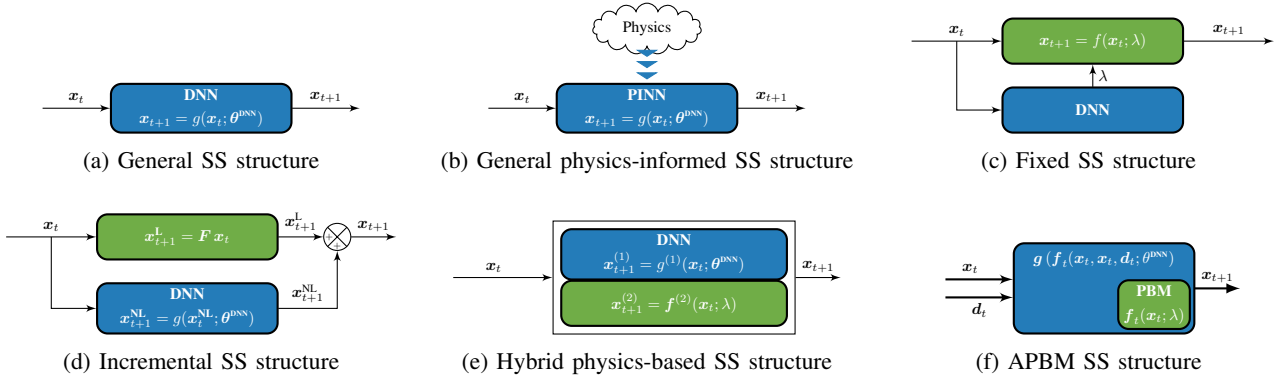
\begin{figure*}
  \setlength{\belowcaptionskip}{1\baselineskip}
\centering
\subcaptionbox{General \ac{ss} structure\label{fig:gsss}}[.32\linewidth]{%
\begin{tikzpicture}[scale=0.6, every node/.style={transform shape},>=latex']
  \node[main,fill=myBlue,align=center] (NN) {\ac{dnn}\\[1mm]$\myVec{x}_{t+1} = g(\myVec{x}_{t};\myVec{\theta}^\dnn)$};
  \coordinate[left = 15mm of NN] (input);
  \coordinate[right = 15mm of NN] (output);
  \draw[->] (input) -- node[above] {$\myVec{x}_{t}$} (NN);
  \draw[->] (NN) -- node[above] {$\myVec{x}_{t+1}$} (output);
  \end{tikzpicture}
  }
\subcaptionbox{General physics-informed \ac{ss} structure\label{fig:gpim}}[.32\linewidth]{%
\begin{tikzpicture}[scale=0.6, every node/.style={transform shape},>=latex']
  \node[main,fill=myBlue,align=center] (NN) {PINN\\[1mm]$\myVec{x}_{t+1} = g(\myVec{x}_{t};\myVec{\theta}^\dnn)$};
  \node[cloud, cloud puffs=15, draw, minimum width=2cm,aspect=3,above=0.8cm of NN] (phys) {Physics};
  \coordinate[left = 15mm of NN] (input);
  \coordinate[right = 15mm of NN] (output);
  \draw[->] (input) -- node[above] {$\myVec{x}_{t}$} (NN);
  \draw[->] (NN) -- node[above] {$\myVec{x}_{t+1}$} (output);
 \draw[draw=myBlue,fill=myBlue,decorate,decoration={shape width=.08cm, shape height=.25cm,
    shape=isosceles triangle, shape sep=.15cm,
    shape backgrounds}] ($(phys.south)+(0,-1mm)$)--(NN.north);
  \end{tikzpicture}
  }
\subcaptionbox{Fixed \ac{ss} structure
\label{fig:fmsd}}[.32\linewidth]{%
\begin{tikzpicture}[scale=0.6, every node/.style={transform shape},>=latex']
  \node[main,fill=myGreen] (main) {$\myVec{x}_{t+1}=f(\myVec{x}_{t};\lambda)$};
  \node[main,fill=myBlue,below=5mm of main] (NN) {\ac{dnn}};
  \coordinate[left=2cm of main] (input);
  \coordinate[right=2cm of main] (output);
  \draw[->] (input) -- node[above] {$\myVec{x}_{t}$} (main);
  \draw[->] (main) -- node[above] {$\myVec{x}_{t+1}$} (output);
  \draw[->] ($(input)!0.5!(main.west)$) |- (NN);
  \draw[->] (NN) -- node[right] {$\lambda$} (main);
  \end{tikzpicture}
  }
  \subcaptionbox{Incremental \ac{ss} structure\label{fig:isssd}}[.32\linewidth]{%
\begin{tikzpicture}[scale=0.6, every node/.style={transform shape},>=latex']
  \node[main,fill=myGreen] (main) {$\myVec{x}^\text{L}_{t+1}=\myVec{F}\,\myVec{x}_{t}$};
  \node[main,fill=myBlue,align=center,below=5mm of main] (NN) {\ac{dnn}\\[1mm]$\myVec{x}^\text{NL}_{t+1} = g(\myVec{x}^\text{NL}_{t};\myVec{\theta}^\dnn)$};
  \node [draw,circle, node distance=1cm,minimum size=0.6cm, right = of main] (sum) {};
  \draw (sum.north east) -- (sum.south west) (sum.north west) -- (sum.south east);
  \node[left=-1pt] at (sum.center){\tiny $+$};
  \node[below] at (sum.center){\tiny $+$};

  \coordinate[left = 2cm of main] (input);
  \coordinate[right = 1cm of sum] (output);
  \draw[->] (input) -- node[above] (xk) {$\myVec{x}_{t}$} (main);
  \draw[->] (xk) |- (NN);
  \draw[->] (main) -- node[above] {$\myVec{x}^\text{L}_{t+1}$}(sum);
  \draw[->] (NN) -| node[above,near start] {$\myVec{x}^\text{NL}_{t+1}$} (sum);
  \draw[->] (sum) -- node[above] {$\myVec{x}_{t+1}$} (output);
  \end{tikzpicture}
  }
\subcaptionbox{Hybrid physics-based \ac{ss} structure\label{fig:hpbmd}}[.32\linewidth]{
\begin{tikzpicture}[scale=0.6, every node/.style={transform shape},>=latex']
      \node[main,fill=myBlue,minimum height=10mm,minimum width =5cm,align=center] (NN) {\ac{dnn}\\[1mm]$\myVec{x}^{(1)}_{t+1} = g^{(1)}(\myVec{x}_{t};\myVec{\theta}^\dnn)$};
      \node[main,fill=myGreen,anchor=north,minimum height=10mm,minimum width=5cm] (main) at (NN.south) {$\myVec{x}^{(2)}_{t+1}=\myVec{f}^{(2)}(\myVec{x}_{t};\lambda)$};
      \node[draw,fit=(NN) (main), inner sep=1mm] (both) {};
      \coordinate[left=22mm of both] (input);
      \coordinate[right = 1cm of both] (output);
      \draw[->] (input) -- node[above,shift={(-0.2,0)}] (xk) {$\myVec{x}_{t}$} (both);
      \draw[->] (both) -- node[above] {$\myVec{x}_{t+1}$} (output);
      \end{tikzpicture}
      }
\subcaptionbox{APBM \ac{ss} structure
\label{fig:apbm_scheme}}[.32\linewidth]{\begin{tikzpicture}[thick,scale=0.6, every node/.style={transform shape},>=latex']
  \node[main,fill=myBlue,minimum height=20mm,minimum width =4cm,text depth = 14mm] (NN) {$\myVec{g}\left(\myVec{f}_t(\myVec{x}_{t}),\myVec{x}_{t},\myVec{d}_{t};\theta^\dnn\right)$};
  \node[main,minimum width=14mm,fill=myGreen,anchor=south east,shift={(-1mm,1mm)},align=center] (main) at (NN.south east) {PBM\\[1mm]$\myVec{f}_t(\myVec{x}_{t};\lambda)$};
\coordinate[left=15mm of NN,yshift=2mm] (input1);
  \coordinate[below = 4mm of input1] (input2);
  \coordinate[right = 1cm of NN] (output);
  \draw[->] (input1) -- node[above,shift={(-0.2,0)}] (xk) {$\myVec{x}_{t}$} (input1-|NN.west);  
          \draw[->] (input2) -- node[below] {$\myVec{d}_{t}$} (input2-|NN.west);
  \draw[-latex] (NN) -- node[above] {$\myVec{x}_{t+1}$} (output);
\end{tikzpicture}
}
\caption{Structures for SS-oriented learning.}
\label{fig:ModelStructures}
\end{figure*}

The approaches to data-augmented modeling of the state equation can be classified in terms of the \ac{ss} components that the \ac{dnn} characterizes or is embedded in.
In the general \ac{ss} structures (see Fig.~\ref{fig:gsss}), the whole model of state dynamics is identified from data and replaced by an~\ac{dnn}\footnote{A structure similar to the general one can be considered for the continuous-time models, where the \ac{dnn} represents the function in the differential state equation, and then the Euler discretization is used to obtain a discrete-time model used for the estimation.
}~\eqref{eq:ddModelState}.
In~\cite{suykens1995nonlinear}, \ac{fc} \acp{dnn} were used to represent the function $f_t(\cdot)$, and in \cite{masti2021learning}, the authors use predictive partial autoencoders to find the state and its mapping to the measurement.
In~\cite{gorji2008identification}, \ac{fc} \ac{dnn}s trained by the expectation-maximization algorithm were used to represent both functions.  
A comparison of the performance of several general \ac{ss} structures was presented in~\cite{gedon2021deep}, considering \ac{lstm}, \ac{gru}, and \acp{vae}.
The authors assumed Gaussian likelihood and combined the \ac{rnn} and \ac{vae} into a deep state space model.
Stochastic \acp{rnn} were introduced for system identification in~\cite{fraccaro2016sequential} to account for stochastic properties of the \ac{ss} model. A similar approach follows \acp{pinn}~\cite{arnold2021state}, modeling the state equation by a \ac{dnn} for which training is constrained by the physics describing the \ac{pbm} (see Fig.~\ref{fig:gpim}).


Another approach to model state equation is to consider a fixed structure such as a linear parameter-varying model~\eqref{eq:LinssModelState} and use the \acp{dnn} to represent the matrices~\cite{bao2020identification} 
(see Fig.~\ref{fig:fmsd}). Such an approach has also been adopted in
\cite{xu2021ekfnet}, where the covariance matrices $\myMat{Q}$ and $\myMat{R}$ were provided by a~\ac{dnn}.


A similar idea was elaborated in~\cite{schoukens2021improved} where an incremental \ac{ss} structure was proposed to separate $f_t(\cdot)$  into linear and nonlinear parts and use a \ac{dnn} to represent the nonlinear part (see Fig.~\ref{fig:isssd}).
In~\cite{forgione2020model}, a hybrid model (see Fig.~\ref{fig:hpbmd}) was proposed, which combines the \ac{pbm} for some state elements and the \ac{dnn} for the rest. It leverages the fact that the dynamic of some states is precisely known (e.g., the position derivative is velocity). 

The data-augmented modeling represented by the \textit{\ac{apbm}} combines the physics- and data-driven components into a single model, which reads
\begin{align}
     \myVec{x}_{t+1} &= g\left(f_t(\myVec{x}_{t}), \myVec{x}_{t}, \myVec{d}_{t};\myVec{\theta}_t^\apbm\right) + \myVec{v}_{t}^\apbm, \label{eq:apbModelState}
\end{align}
where  $\myVec{v}_t^\apbm$ is a process noise, $\myVec{d}_{t}$ represents additional data\footnote{These denote data related to the system available to the user but not used in the \ac{pbm} as additional inputs to avoid overly
complex models (e.g., ionospheric models, weather forecasts).}, the function $g(\cdot)$ is aware of the \ac{pbm} part, and the vector $\theta_t^\apbm$ is designed to minimize the discrepancy between outputs of \eqref{eq:apbModelState} and the \ac{tm}.
The \ac{apbm} compensates the PBM structure and parameter mismodelling using information extracted from available data. Consequently, the \ac{apbm} preserves the physical meaning of the state components and exploits actual system behavior dependencies, which were ignored in the \ac{pbm} design. 
An example of this modeling versatility was shown in \cite{TangAPBMHOM2024}, where \acp{apbm} were used to filter a high-order Markov process without the need for order selection.
One important characteristic of the APBM formulation~\eqref{eq:apbModelState} is the flexibility to cope with the non-stationarity of the model over time or space ($\myVec{\theta}^\apbm_{t-\tau}\neq\myVec{\theta}^\apbm_{t}$), which requires adaptive estimation strategies~\cite{imbiriba2022hybrid}.

The general \ac{apbm} structure \eqref{eq:apbModelState} can be simplified into an additive form with explicitly controlled contribution of the data-based component (see box entitled {\em `\ac{apbm} with Controlled Additive Structure'} on Page~\pageref{Box:apbm}). Bounding this contribution 
is an essential feature of the \ac{apbm}, which prevents the data-based component from overruling the \ac{pbm} component contribution and, thus, preserving \ac{apbm} explainability.
The \ac{apbm} state dynamics model \eqref{eq:apbModelState} together with the observation model (any of \eqref{eq:ssModelObs}, \eqref{eq:LinssModelObs}, \eqref{eq:NLssModelObs}) can directly be used for joint state $\myVec{x}_{t}$ and parameter $\myVec{\theta}_t^\apbm$ estimation by a regular state estimator such as the EKF or the UKF \cite{sarkka2023bayesian}.


{\bf{Training APBMs:}}
Training APBMs can be performed under different paradigms depending on data availability, architecture, and assumptions regarding the stationarity of the system. 
Although the learning strategy can be supervised when $\mySet{D}_{\rm s}$ is available, we will focus on the more common problem in the Bayesian filtering literature, where only noisy observations, $\mySet{D}_{\rm u}$, are accessible, setting up an unsupervised learning scenario.
In this context, parameter estimation can be achieved by $(i)$ obtaining and maximizing the marginal posterior $p(\myVec{\theta}|\myVec{y}_{1:T})$, often using the energy function ($\varphi_T(\myVec{\theta}) = -\log p(\myVec{y}_{1:T}| \myVec{\theta}) - \log p(\myVec{\theta})$)
~\cite{sarkka2023bayesian};
$(ii)$ the joint posterior $p(\myVec{x}_t,\myVec{\theta}|\myVec{y}_{1:T})$
~\cite{imbiriba2022hybrid};
or $(iii)$ obtaining a point estimate $\hat{\myVec{\theta}}$ through deterministic optimization strategies often aiming at maximizing the variational lower bound of the log-likelihood $p(\myVec{y}_{1:T};\myVec{\theta})$~\cite{krishnan2017structured}.

In~\cite{imbiriba2022hybrid, imbiriba2023augmented}, the authors opted for a state-augmentation approach aiming at obtaining the joint posterior distribution $p(\myVec{x}_t,\myVec{\theta}_t|\myVec{y}_{1:T})$ through Bayesian filtering recursion. For such, system states are augmented with the APBM parameters:
\begin{align}
    \begin{bmatrix}
    \myVec{x}_{t+1}\\
    \myVec{\theta}^\apbm_{t+1}
    \end{bmatrix} = 
    \begin{bmatrix}
    g\left(f_t(\myVec{x}_{t}),\myVec{x}_{t}, \myVec{d}_{t};\myVec{\theta}^\apbm_t\right)\\
    \myVec{\theta}^\apbm_{t}
    \end{bmatrix} + 
    \begin{bmatrix}
    \myVec{v}_{t}^{\myVec{x}} \\
    \myVec{v}_{t}^{\myVec{\theta}}
    \end{bmatrix},
     \label{eq:apbm_dynamics}
\end{align}
where a near-constant state transition process is introduced for $\myVec{\theta}^\apbm$, allowing one to cast the APBM learning as a filtering problem.  $\myVec{v}_{t}^{\myVec{\theta}}$ is a ``small'' noise and is introduced to avoid numerical issues
~\cite{sarkka2023bayesian}.
Note that such noise can also allow $\myVec{\theta}^\apbm_t$ to drift over time and eventually evolve if the model is time-varying.  

Another important component regarding the APBM learning process is the need for a control mechanism to prevent the DNN component from overpowering the PBM. The prevention can be achieved by imposing appropriate constraints over $\myVec{\theta}^\apbm$. A Bayesian filtering  
 based approach is to introduce pseudo-observation equations into the observation model. 
 Let $\Bar{\myVec{\theta}}$ be a vector such that $g\left(f_t(\myVec{x}_{t}),\myVec{x}_{t},\myVec{d}_{t};\myVec{\theta}^\apbm_t = \Bar{\myVec{\theta}}\right) = f_t(\myVec{x}_{t})$, then the observation model in~\eqref{eq:NLssModelObs} can be augmented as:
\begin{align}
    \begin{bmatrix}
    \myVec{y}_{t}\\
    \Bar{\myVec{\theta}}
    \end{bmatrix} = 
    \begin{bmatrix}
    h_t(\myVec{x}_{t})\\
    \myVec{\theta}^\apbm_t
    \end{bmatrix} + 
    \begin{bmatrix}
    \myVec{w}_{t}\\
    \myVec{w}_{t}^{\myVec{\theta}}
    \end{bmatrix},
    \label{eq:apbm_measurement}
\end{align}
where the bottom equation acts as a constraint, forcing $\myVec{\theta}^\apbm_t$ to be in the vicinity of $\Bar{\myVec{\theta}}$; and the distribution of $\myVec{w}_{t}^{\myVec{\theta}}$ governs the trade-off between regularization and data fit.
In the case where the noise terms in the SS model defined in Equations~\eqref{eq:apbm_dynamics} and~\eqref{eq:apbm_measurement}  are Gaussian, $\text{cov}\{\myVec{\omega}_t\}=\myVec{R}$ and $\text{cov}\{\myVec{\omega}^\myVec{\theta}_t\}=\eta^{-1}\myVec{I}$, the Bayesian filtering solution can be cast as a sequence of minimization problems for every time step $t$:
\begin{align}
    (\hat{\myVec{x}}_t, \hat{\myVec{\theta}}^\apbm_t) &= \mathop{\arg\min}_{(\myVec{x}_t, \myVec{\theta}_t)}\|\myVec{y}_t - h_t(\myVec{x}_t)\|^2_{\myVec{R}^{-1}} + \eta \| \bar{{\myVec{\theta}}} - {\myVec{\theta}_t}\|^2  \nonumber\\
    &+ \|\myVec{x}_t - g\left(f_t(\myVec{x}_{t-1}),\myVec{x}_{t-1}, \myVec{d}_{t-1};\hat{\myVec{\theta}}^\apbm_{t-1}\right)\|^2_{[\hat{\myVec{P}}_{t|t-1}^x]^{-1}} 
    + \|\myVec{\theta}_t - \hat{\myVec{\theta}}^\apbm_{t-1}\|^2_{[\hat{\myVec{P}}_{t|t-1}^\myVec{\theta}]^{-1}}, \label{eq:online_costfunc}
\end{align}
where $\eta\geq 0$ controls the strictness of the regularization added by the pseudo-observation equation in~\eqref{eq:apbm_measurement}.

{\textit{Offline vs Online Training of APBMs:}}
One interesting aspect of leveraging the Bayesian filtering approach for joint parameter and state estimation lies in its equivalence to a second-order Newton method~\cite{humpherys2012fresh}, leading to fast convergence and making it suitable for both online and offline training. The training methodology seamlessly allows for either offline, online, or both (offline-online) training. 
In the offline scenario, the filtering with the augmented model can be performed over the multiple sequences (and multiple epochs) in $\mySet{D}_{\rm u}$ if system states, $\myVec{x}_0$, are properly initialized for every data sequence in $\mySet{D}_{\rm u}$~\cite{Bemporad_kftraining_2023}. Once training is completed, a standard Bayesian filtering approach can be used to update only the system states, $\myVec{x}_t$, over time while keeping the APBM parameters fixed. In the online setting, the training approach becomes a conventional filtering problem that continuously adapts both states and model parameters over time, starting from some initial condition $\hat{\myVec{x}}_0$, $\hat{\myVec{\theta}}_0^\apbm$. Again, the fast convergence of the Bayesian filter allows for quick reaction of the data-driven component over time, correcting the PBM to adapt to new conditions and improving the state estimation performance. This feature is extremely important when dealing with non-stationary dynamics that  change over time. Both strategies can be combined, where the offline training solution is used as the initial condition for the online procedure, which, in turn, keeps updating both states and parameters during the test.

\begin{tcolorbox}[float*=t,
    width=\boxWidth,
	toprule = 0mm,
	bottomrule = 0mm,
	leftrule = 0mm,
	rightrule = 0mm,
	arc = 0mm,
	colframe = myblue,
	colback = mypurple,
	fonttitle = \sffamily\bfseries\large,
	title = \ac{apbm} with Controlled Additive Structure]	
	\label{Box:apbm}
\begin{wrapfigure}{r}{7cm} 
    \centering
    \includegraphics[width=7cm]{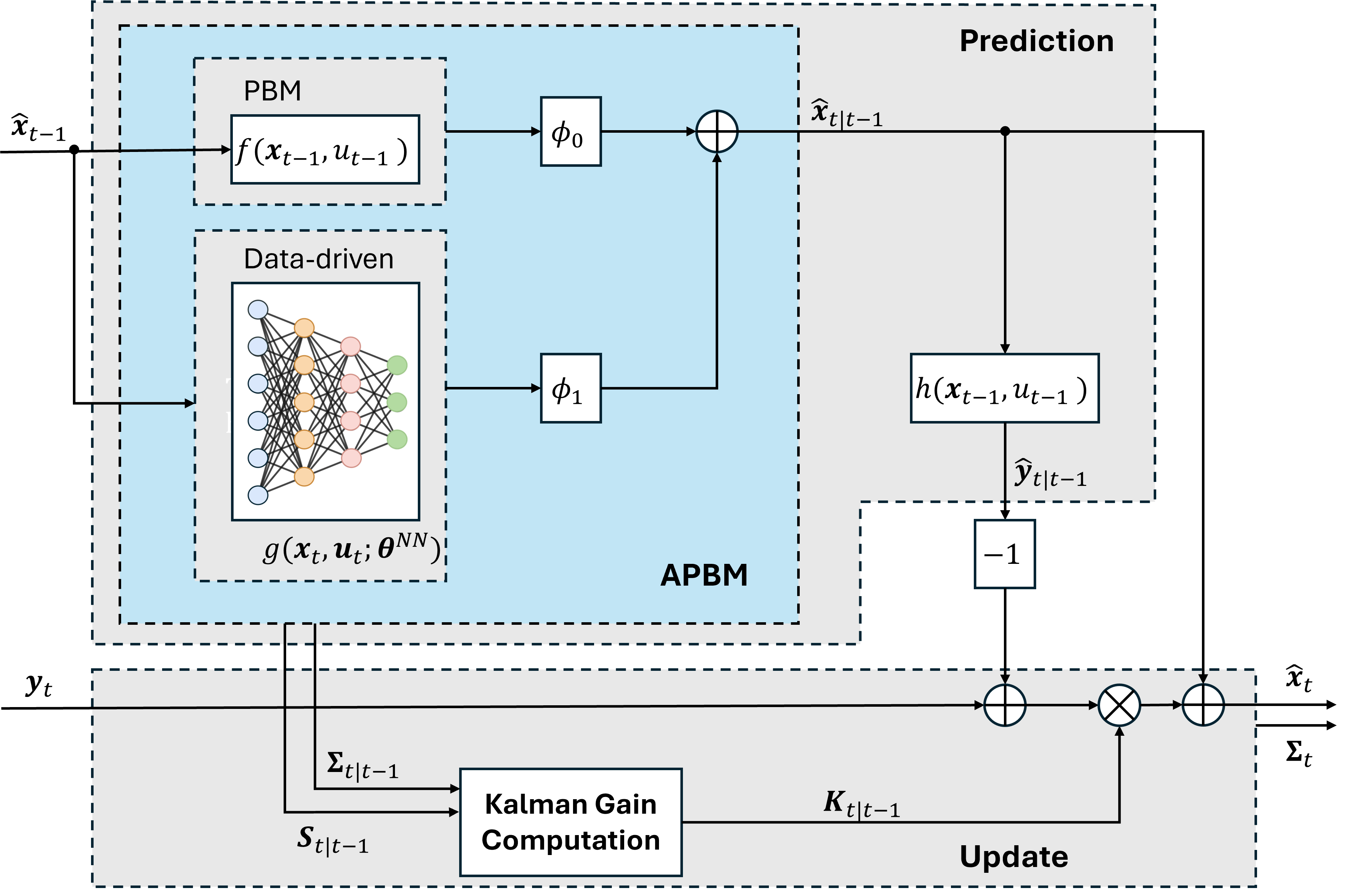} 
    \caption{\ac{apbm} with additive structure.} 
    \label{fig:apbm}
\end{wrapfigure}

The general \ac{apbm} in \eqref{eq:apbModelState} can be particularized as an additive augmentation of the \ac{pbm} \cite{imbiriba2023augmented}
\begin{align}
     \myVec{x}_{t+1} &= \phi_0f_t(\myVec{x}_{t})+\phi_1g(\myVec{x}_{t};\myVec{\theta}^\dnn) + \myVec{v}_{t}^\apbm, \label{eq:apbModelState_additive}
\end{align}
where the \ac{apbm} parameters $\myVec{\theta}^\apbm=[\phi_0,\phi_1,\myVec{\theta}^\dnn]$ may be estimated offline or online, with supervised or unsupervised training. 
To ensure that the data-based component does not overrule the \ac{pbm} component, we can use \textit{constrained} state estimation algorithms where the \ac{dnn} parameters are \textit{encouraged} to be close to a nominal value $\bar{\bm{\theta}}$ that keeps the \ac{dnn} contribution limited. 
State estimation  in this case is  interpreted as a regularized optimization problem
\begin{equation*}
    (\hat{\myVec{x}}_{1:t}, \bm{\theta}) = \mathop{\arg\min}_{(\myVec{x}_{1:t},\bm{\theta})} \mathcal{L}_{\mySet{D}_{\rm u}} (\myVec{y}_{1:t},{\color{NewColor}\bar{\myVec{y}}_{1:t}}) + \eta \mathcal{R}(\myVec{x}_{1:t},\bm{\theta}),
\end{equation*}
where $\mathcal{L}_{\mySet{D}_{\rm u}}$ and $\cal{R}$ are cost and regularization functionals, 
\color{NewColor}
$\eta\in\mathbb{R}_+$ is a regularization parameter governing the trade-off between model fit and regularization, and $\bar{\myVec{y}}_{t}\triangleq h_t(\myVec{x}_t)$. 
\color{black}
In the context of Kalman filtering, the minimized cost function is the MSE of the state given observations and dynamics~\eqref{eq:apbModelState_additive} model such as~\eqref{eq:online_costfunc}. The \ac{dnn} can be effectively controlled through the regularization term, for which several implementation options are possible \cite{imbiriba2023augmented}. The concept of the KF-based estimation of the APBM with controlled additive structure is illustrated in Fig. \ref{fig:apbm}.
\end{tcolorbox}

{\bf Discussion:} 
The SS-oriented AI-augmented KF design relies on augmenting the physics-based component, namely the ``deterministic'' part of the state equation with a data component. As seen in the box entitled {\em `\ac{apbm} with Controlled Additive Structure'}, a solution might lead to an additive \ac{apbm} state equation \eqref{eq:apbModelState_additive}. The data component is designed to extract a \textit{time-correlated} component of the discrepancy between the pure \ac{pbm} \eqref{eq:NLssModelState}  and the \ac{tm}, that is usually embraced by the overbounded process noise $\myVec{v}_{t}$. As a consequence, the \ac{apbm} process noise $\myVec{v}_{t}^\apbm$ has different statistical properties from the \ac{pbm} $\myVec{v}_{t}$ and for optimal performance of the \ac{kf}, the noise properties have to be identified. In \cite{DuStKoTaImCl:24}, the noise properties identification of the \ac{apbm} was discussed and illustrated using correlation and maximum likelihood methods. 
Utilizing the identified process noise covariance matrix in the \ac{kf} led to significant improvement of the estimate consistency.

Following the terminology used in system identification \cite{Lj:08}, the \ac{pbm} augmentation with \ac{dnn} belongs into \textit{block-oriented ``slate-gray'' models}. The idea of block-oriented models is ``to build up structures from simple building blocks'' \cite{Lj:08}, which allows physical insight and data-oriented flexible complement. Note that besides the \ac{dnn}, the \ac{pbm} can be augmented with the \acl{narma} model as in \cite{LiZhLi:21}, where the data component is used for drift compensation. The  \ac{apbm} concept can also be used for deterministic models with a completely measured state, for characterizing unmodelled aspects of the dynamics. 

\section{Comparative Study}
\label{sec:Comparative}
The previous sections presented a set of design approaches and concrete algorithms fusing Kalman-type filtering and \ac{ai} techniques. While all these methods tackle the same task of state estimation in dynamic systems, they notably vary in their strengths, requirements, and implications. To highlight the interplay between the approaches mentioned above and provide an understanding of how practitioners should prioritize one approach over the others, we next provide a comparative study. We divide this study into two parts: we first present a {\em qualitative comparison}, which pinpoints the conceptual differences between the approaches in light of the identified desired properties \ref{itm:Optimality}-\ref{itm:complexity} and challenges \ref{itm:FuncAccuracy}-\ref{itm:InferenceSpeed}. We then detail a {\em quantitative study} for a representative scenario of tracking in challenging dynamics to capture the regimes in which each approach is expected to be preferable. 

\subsection{Qualitative Comparison}
\label{ssec:Qualitative}
Here, we discuss the relationship and  gains of the different approaches over each other in terms of  figures of merit that are not quantifiable in the same sense as estimation performance on a given test-bed is. Based on the identified desired properties \ref{itm:Optimality}-\ref{itm:complexity} and challenges \ref{itm:FuncAccuracy}-\ref{itm:InferenceSpeed} of conventional Kalman-type algorithms, we focus on qualitative comparison in terms of domain knowledge, interpretability, uncertainty extraction, adaptability, target family of \ac{ss} models, and learning framework. The comparison detailed below is summarized in Table~\ref{Tbl:Comparison}.

\begin{table*}
\centering 
\setlength{\tabcolsep}{2pt} 
\renewcommand{\arraystretch}{1.5}
{\scriptsize
\begin{tabular}{|p{1.1cm} |p{1.5cm}|p{2.5cm}|p{2.2cm}|p{2.1cm}|p{2.1cm}|p{2.1cm}|p{2.5cm}|}
 \hline
 \multicolumn{2}{|c|}{{\bf Approach}} & {\bf \ac{ss} Knowledge} & {\bf Interpretability} & {\bf Uncertainty} & {\bf Adaptability} & {\bf \ac{ss} Model} & {\bf Learning} \\ \hline \hline      
 
 \multirow{2}{1.2cm}{\centering End-to-end \ac{dnn}}  
       & Generic
       & Fully model-agnostic\cellcolor[HTML]{AAFDB4}          & Black-box\cellcolor[HTML]{FF9595}         & Only outputs state\cellcolor[HTML]{FF9595}         &  Not adaptive\cellcolor[HTML]{FF9595}     &  Can be non-linear and non-Gaussian\cellcolor[HTML]{AAFDB4}      & Supervised learning from large data sets\cellcolor[HTML]{FF9595}                   \\ \cline{2-8} 
     & \ac{kf}-inspired (e.g., \ac{rkn} \cite{becker2019recurrent})
       & Fully model-agnostic\cellcolor[HTML]{AAFDB4}         & Connection of black-box \acp{dnn}\cellcolor[HTML]{FF9595}         & Estimates both state and uncertainty\cellcolor[HTML]{AAFDB4}        &   Not adaptive\cellcolor[HTML]{FF9595}     &   Can be non-linear and non-Gaussian\cellcolor[HTML]{AAFDB4}   & Supervised learning from large data sets\cellcolor[HTML]{FF9595}    \\ \specialrule{.2em}{.1em}{.1em}
\multirow{2}{1.2cm}{\centering External \ac{dnn}}  
       & Learned pre-process
       & Requires state evolution\cellcolor[HTML]{FFEAAD}         & \cellcolor[HTML]{FFEAAD} Input transformation is black-box, while tracking based on the latent representation is fully interpretable        & Tracks both state and uncertainty\cellcolor[HTML]{AAFDB4}         &  Adapts to known variations only in the state evolution\cellcolor[HTML]{FFEAAD}       &  State evolution should be simple (preferably linear) and Gaussian\cellcolor[HTML]{FFEAAD} & Preferably supervised for end-to-end learning, though \ac{dnn} can also be trained as a feature extractor, possibly unsupervised \cellcolor[HTML]{FFEAAD}                     \\ \cline{2-8} 
     & Learned correction (e.g., \cite{satorras2019combining})
       &  Requires (estimated) \ac{ss} representation\cellcolor[HTML]{FF9595}         & Highly interpretable \cellcolor[HTML]{AAFDB4}          & Tracks both state and uncertainty\cellcolor[HTML]{AAFDB4}            &  Designed for a specific \ac{ss} model \cellcolor[HTML]{FF9595}    &   Should be Gaussian with simple non-linearities\cellcolor[HTML]{FF9595}   & Supervised learning from moderate data sets\cellcolor[HTML]{FF9595}    \\ \specialrule{.2em}{.1em}{.1em}
\multirow{2}{1.2cm}{\centering Integrated \ac{dnn}}  
       & Learned \ac{kg} (e.g., KalmanNet \cite{revach2022kalmannet})
       & Requires (estimates) of $h(\cdot)$ and $f(\cdot)$\cellcolor[HTML]{FFEAAD}          & \cellcolor[HTML]{FFEAAD} Processing follows the same pipeline as standard \ac{ekf}, with \ac{kg} computation being black-box        & Can be extracted in some scenarios~\cite{Dahan2024uncertainty}\cellcolor[HTML]{FFEAAD}          &   Requires re-training~\cite{revach2021unsupervised} or hypernetworks~\cite{ni2023adaptive} for adaptivity\cellcolor[HTML]{FF9595}    &  Can be non-linear and non-Gaussian\cellcolor[HTML]{AAFDB4}        & Preferably supervised for end-to-end learning, can also be trained unsupervised with additional domain knowledge~\cite{revach2021unsupervised} \cellcolor[HTML]{FFEAAD}                 \\ \cline{2-8} 
     & Learned State Estimation (e.g., DANSE \cite{ghosh2023danse})
       & \cellcolor[HTML]{FFEAAD} Requires SS observation model  and does not require SS state evolution model       & \cellcolor[HTML]{FFEAAD} Posterior update is interpretable and resembles that of SS-oriented KF, while state prediction is black-box        & \cellcolor[HTML]{AAFDB4} Tracks both state and uncertainty        & \cellcolor[HTML]{FF9595} Requires re-training for adaptivity    &  \cellcolor[HTML]{FFEAAD}State evolution model can be non-linear, observation model should be linear, Gaussian  & Unsupervised learning\cellcolor[HTML]{AAFDB4}  \\ \specialrule{.2em}{.1em}{.1em}
\multirow{2}{1.2cm}{\centering \ac{ss}-Oriented \ac{dnn}}  
       & DDM (e.g., fully data-driven model or PINN \cite{raissi2019physics})
        &  Fully model-agnostic (DNN) or some knowledge of the model required (PINN). The observation model is assumed to be known. 
       \cellcolor[HTML]{FFEAAD}         & The state evolution is black-box, while the filter operation is interpretable 
       \cellcolor[HTML]{FFEAAD}        & Tracks both state and uncertainty
       \cellcolor[HTML]{AAFDB4}        &  Not adaptive\cellcolor[HTML]{FF9595}    &  Can be non-linear and non-Gaussian\cellcolor[HTML]{AAFDB4}
       & Supervised learning  \cellcolor[HTML]{FF9595}             \\ \cline{2-8} 
       & Parameter learning (e.g., \cite{bao2020identification})
       & Requires models of $h(\cdot)$ and $f(\cdot)$\cellcolor[HTML]{FFEAAD}         & Fully interpretable\cellcolor[HTML]{AAFDB4}        & \cellcolor[HTML]{AAFDB4} Tracks both state and uncertainty         &  Not adaptive\cellcolor[HTML]{FF9595}    &  Can be non-linear and non-Gaussian\cellcolor[HTML]{AAFDB4}        & Unsupervised learning   \cellcolor[HTML]{AAFDB4}            \\ \cline{2-8} 
     & APBM (e.g., \cite{imbiriba2023augmented})
       & Requires models of $h(\cdot)$ and $f(\cdot)$\cellcolor[HTML]{FFEAAD}         & Highly interpretable\cellcolor[HTML]{AAFDB4}        & \cellcolor[HTML]{AAFDB4} Tracks both state and uncertainty        &  Fully adaptive\cellcolor[HTML]{AAFDB4}    &  Can be non-linear and non-Gaussian\cellcolor[HTML]{AAFDB4}  & Unsupervised learning \cellcolor[HTML]{AAFDB4}  \\ \specialrule{.2em}{.1em}{.1em}
  \centering  Model-Based & Kalman-type filters (e.g., \ac{ekf})  & Requires (accurate) \ac{ss} representation\cellcolor[HTML]{FF9595}          & Fully interpretable\cellcolor[HTML]{AAFDB4}          & Tracks both state and uncertainty\cellcolor[HTML]{AAFDB4}        &  Fully adaptive\cellcolor[HTML]{AAFDB4}     &  Should be Gaussian with mild non-linearities\cellcolor[HTML]{FF9595}  &  Fully model-based\cellcolor[HTML]{AAFDB4}    \\ \hline   

\end{tabular}
} 

    \caption{Qualitative comparison between the considered approaches.}
    \label{Tbl:Comparison}
\end{table*}

\subsubsection{Domain Knowledge} 
The presented methodologies substantially vary in the required knowledge of the \ac{ss} model, corresponding to challenges \ref{itm:FuncAccuracy} and \ref{itm:Stochasticity}. The extreme cases are those of fully model-based filters (such as the  \ac{kf} and its variants) that require full and accurate knowledge of the \ac{ss} model, and those of end-to-end \acp{dnn}, that are  model-agnostic (yet do not incorporate  characterization when  provided). 

Among the methodologies representing \ac{ai}-augmented \acp{kf}, the usage of an external \ac{dnn} to classic Kalman-type tracking typically requires the same level of domain knowledge needed to apply its classic counterpart. When the \ac{dnn} is applied in parallel as a learned correction term, then the complete \ac{ss} model should be known, while applying a learned pre-processing module can compensate for unknown and intractable observation modeling. Settings in which one has full knowledge of the observations model but does not know the state evolution model are the focus of methods based on learned state estimation and the \ac{ss}-oriented \ac{ddm} and \ac{pinn}. Knowledge (though possibly an approximated one, as in \ref{itm:FuncAccuracy}) of the functions $f(\cdot)$ and $h(\cdot)$ is required by techniques that learn the \ac{kg}, while some modeling of these functions is needed by \ac{ss}-oriented approaches based on parameter learning and \ac{apbm}.

\subsubsection{Interpretability} 
By interpretability  we refer to the ability to explain the operation of each computation and associate its internal features with concrete meaning, as in \ref{itm:interpretability}. This high level of interpretability is naturally provided by purely model-based \ac{kf}-type algorithms, as well as by \ac{ss}-oriented \ac{ai} techniques that learn the parameters of a pre-determined parametric \ac{ss} model~\cite{bao2020identification}, similarly to conventional system identification. On the contrary, algorithms employing end-to-end \acp{dnn} for state estimation are essentially black-box methods.

The level of interpretability varies when considering algorithms that augment classic algorithms with \acp{dnn}, such that some of the computations are based on principled statistical models, while some are based on black-box data-driven pipelines. For instance, the internal features of task-oriented designs with integrated \acp{dnn} are exactly those of standard Kalman-type algorithms, while some of the internal computations -- such as the computation of the \ac{kg} in KalmanNet or the propagation of the prior state moments in DANSE -- are carried out by black-box \acp{dnn}. A higher level of interpretability is provided when the \ac{dnn} is applied in parallel to a fully model-based and interpretable algorithm for learned correction, as in~\cite{satorras2019combining}.

\subsubsection{Uncertainty} 
The ability to provide faithful {\em uncertainty} measures (as in \ref{itm:uncertainty}) arises in Kalman-type algorithms that track of the error (posterior) covariance matrix. As the posterior update is an integral aspect of the \ac{kf} {\em update} step, any algorithm that fully implements this computation provides uncertainty measures along with its state estimate. This property is exhibited by fully model-based Kalman-type filters, as well as by some \ac{ai}-aided \acp{kf}. Specifically, both \ac{ss}-oriented designs, which implement conventional filtering on top of a learned state evolution model, as well as task-oriented architecture that augments the prior state prediction 
or utilize \acp{dnn} external to model-based filters provide uncertainty in the update computation.

Among \ac{ai}-aided algorithms that do not  preserve the posterior covariance propagation of \acp{kf}, the ability to provide uncertainty varies. Generic end-to-end \acp{dnn} that track only the state do not offer such measures, while \ac{kf}-inspired architectures are designed to output error covariances via their update \acp{dnn}. \ac{ai}-aided \acp{kf} with learned \ac{kg} (such as KalmanNet) are specifically designed to bypass the need to propagate second-order moments such as the posterior covariance. Still, as noted in the box entitled {\em `Uncertainty Extraction from Learned \ac{kg}`} on Page~\pageref{Box:KalmanNet}, in some settings, one can still extract uncertainty measures from their internal \ac{kg} features.

\subsubsection{Adaptability} 
As noted in \ref{itm:Adaptability}, \ac{kf}-type algorithms are adaptable to temporal variations in the statistical model. Assuming that one can identify and characterize the variations, the updated \ac{ss} model parameters are simply substituted into the filter equations. End-to-end \acp{dnn}, in which there is no explicit dependence on the \ac{ss} model parameters, and these are implicitly embedded in the learned weights, adapting the \ac{ss} models that deviate from those observed in training necessitates re-training, i.e., they are not adaptive.

Task-oriented \ac{ai}-aided \acp{kf} are trained in a manner that is entangled with the operation of the augmented algorithm. While algorithms with integrated and external \acp{dnn} typically use the \ac{ss} model parameters in their  processing, their \ac{dnn} modules are fixed, and are suitable for the \ac{ss} models observed in training. Accordingly, adaptations require re-training, where some level of variations can still be coped with using hypernetworks~\cite{ni2023adaptive}. An exception is the usage of external \acp{dnn} for feature extraction, which is typically designed to map complex observations into a simplified observation model and thus are not affected by variations in the state evolution model. Among \ac{ss}-oriented approaches, adaptivity is provided by the design of \ac{apbm}, while alternatives based on, e.g., \ac{ddm} or \ac{pinn} architectures need retraining when the state evolution statistics change.

\subsubsection{\ac{ss} Model Type}
A key distinct property among the presented methodologies lies in the family of {\em \ac{ss} models} for which each algorithm is suitable. As noted in \ref{itm:NonLinear}, fully model-based \ac{kf}-type algorithms are most suitable for Gaussian \ac{ss} models with simple and preferably not highly non-linear state evolution and observation models. On the opposite edge of the spectrum are end-to-end model-agnostic \acp{dnn}, that can be applied in complex and intractable dynamic systems, provided with sufficient data corresponding to that task.

Among task-oriented \ac{ai}-augmented \acp{kf}, architectures employing external \acp{dnn} as learned correction terms are suitable for similar \ac{ss} models as classic \acp{kf}, being geared towards settings where the latter is applicable. External \acp{dnn} that pre-process observations facilitate coping with complex observation models, while integrated \acp{dnn} for learned state estimation overcome complex state evolution models, with each approach requiring the remainder of the \ac{ss} model to be simple (preferably linear), fully known, and Gaussian. Augmenting \acp{kf} with learned \ac{kg} notably enhances the family of applicable \ac{ss} models, not requiring Gaussianity and modeling of any of the noise terms, and learning to cope with non-linear and possibly approximated state evolution and observation functions. The same holds for the reviewed \ac{ss}-oriented methods that can all cope with non-linear and non-Gaussian dynamics.


\subsubsection{Learning Framework}
The learning framework is specific to algorithms that incorporate \ac{ai}, and encapsulates the amount of data and the reliance on its labeling. Accordingly, fully model-based algorithms that rely on given mathematical modeling of the \ac{ss} representation do not involve learning in their formulation, while black-box \acp{dnn} typically require learning from large volumes of labeled data sets.

Designs combining partial domain knowledge with \ac{ai} typically result in a dominant inductive bias that allows training with limited data, compared to black-box  \acp{dnn}. Methods based \ac{ss}-oriented \ac{ddm}/\ac{pinn}, external \ac{dnn}, and learned integrated \ac{kg} \ac{dnn}, typically require labeled data, though for learned pre-processing and for KalmanNet, one can also train unsupervised in some settings. Methods such as DANSE, \ac{ss}-oriented parameter learning, and \ac{apbm} are  designed to be trained based solely on observations, i.e., unsupervised.


%
%
%
\begin{table*}
\begin{center}
\caption{MSE $\dB$ - Lorenz Attractor with Sampling Mismatch.}
 {
\begin{tabular}{|c|c|c|c|c|c|c|}
\hline
\rowcolor{lightgray}
Noise & EKF & PF &
KalmanNet & DANSE & APBM (offline) & APBM (online)
\\
\hline
-0.024 & -6.316  & -5.333 & 
-11.106 & -10.115 & -9.457 & -7.452
\\
$\pm$ 0.049 & $\pm$ 0.135 & $\pm$ 0.136 &
$\pm$ 0.224 & $\pm$ 0.163 & $\pm$ 0.231 & $\pm$ 0.548
\\
\hline
\end{tabular}
\label{tbl:decimation}
}
\end{center}
\vspace{-0.2cm}
\end{table*} 
\subsection{Quantitative Comparison}
\label{ssec:Quantitative}
%
%
To illustrate the performance of the considered state estimation algorithms, we provide a dedicated experimental study\footnote{The source code for all  algorithms and the hyperparameters  can be found online at \url{https://github.com/ShlezingerLab/AI_Aided_KFs}}. We consider the task of tracking the nonlinear movement of a free particle in three-dimensional space $(m=3)$ from noisy position observations. The Lorenz Attractor, a chaotic solution to the Lorenz system of ordinary differential equations, defines the continuous-time state evolution of the particle's trajectory. To enhance the challenge of this tracking case, we introduce additional uncertainty due to a sampling-time mismatch in the observation process. While the underlying ground truth synthetic trajectory is generated using a high-resolution time interval $\left(\Delta\tau=10^{-5}\right)$, the tracking filter can access only noisy observations decimated at a rate of $\frac{1}{2000}$, resulting in a decimated process with $\Delta{t}=0.02$.

To ensure a fair evaluation, we restrict any populated evolution model of the tracking filter, if it exists, to discrete time with $\Delta{t}\geq0.02$. Further details about the Lorenz Attractor evolution model can be found in~\cite{RTSNet_TSP}. The averaged \ac{mse} values and their standard deviation values for filtering $10$ sequences with a length of $T=3000$ time steps are reported in Table~\ref{tbl:decimation}. There, we compare the \ac{mse} achieved by using the noisy observations as state estimate ({\em Noise}); filtering via the model-based \ac{ekf} and \ac{pf}; the \ac{dnn}-integrated KalmanNet and DANSE; and two forms of the \ac{ss}-oriented \ac{apbm}, employing offline and online learning. 
In the comparative performance reported in Table~\ref{tbl:decimation}, the model-based filters, i.e., the \ac{ekf} and \ac{pf} manage to improve upon using noisy observations, but suffer from an error floor due to their sampling mismatch. All considered \ac{ai}-aided filters manage to improve upon this error floor. Specifically, KalmanNet, which is trained in a supervised manner, achieves the best performance, improving by approximately 5 dB in \ac{mse} compared to the model-based algorithms. Among the \ac{ai}-aided filters that are trained from unlabeled data, DANSE achieves the best performance. APBM, which is designed to boost adaptivity, achieves MSE within a small gap of DANSE when trained offline. 

\textcolor{NewColor}{The above quantitative results are evaluated in a controlled experimental setup, which allows to capitalize on the differences in performance arising from the inherent properties of the considered algorithms. In that sense, they  complement the conceptual comparison provided in Table~\ref{Tbl:Comparison}, in revealing the interplay between the reviewed methods for combining deep learning with classic \ac{kf}-type filtering. We note that additional evaluations of these methodologies in application-specific settings and with real-world data are reported in the literature.  Representative examples include the comparison of \acp{kf}, \ac{rnn}, and augmented \acp{kf} for brain-machine interface~\cite{cubillos2025exploringSHORT}; evaluation of \ac{dnn}-aided \acp{kf} for coastline monitoring in~\cite{aspragkathos2023event}; and the tracking results comparing \ac{dnn}-aided \acp{kf} with limited training data in~\cite{chen2025maml}.}

\section{Future Research Directions}
\label{sec:future_research}
  The tutorial-style presentation of designs, algorithms, and experiments of \ac{ai}-aided \ac{kf}s indicate potential gains of proper fusion of model-based tracking algorithms and data-driven deep learning techniques. These, in turn, give rise to several core research directions that can be explored to further unveil the potential, strengths, and prospective use cases of such designs. Exploring these directions is expected to further advance the development and understanding of algorithmic tools to jointly leverage classic \ac{ss} model-based \ac{kf}-type algorithms with emerging deep learning techniques, and preserving the individual strengthens of each approach that has a bearing on the potential of alleviating some of the core challenges shared by a broad range technologies tracking dynamic systems and state estimation. We thus conclude this article with a discussion of some of the open challenges that can serve as future research directions.

\begin{enumerate}

    \item {\em Time-varying \ac{ss} models and adaptation}: A practical state estimation method might need to cater for a scenario where the observation model  \eqref{eq:ssModelObs} is mismatched between training and testing stages, and/or the observation or state evolution functions
    are time-varying without a clear pattern. Typically \ac{ai}-aided \acp{kf} are tied to a fixed observation model, typically not time-varying, and remains same between the the training stage and inference stage. Therefore,  future research may aim to design AI-aided state estimation methods that  adapt to time-varying  \ac{ss} 
    models, mainly for time-varying observation models.  
\item {\em Non-Markovian SS models}: Throughout this article and being consistent with usual practice, the  state evolution model is Markovian  (see \eqref{eq:ssModelState}). Naturally, it is a quest to design methods that can exploit short and long-term memories in state evolution, which means non-Markovian state evolution.  

\item {\em Non-Gaussian SS models}: Most of the directions discussed focus on Gaussian noises in \ac{ss} models. Extensions to non-Gaussian noises will require more complex algorithms with high computational complexity, which is challenging, especially when learning the large-scale \ac{dnn} parameters.
\item {\em Distributed \ac{ai}-aided \acp{kf}}: While there exists considerable research on distributed KFs, such as the work of \cite{DistributedKF_2007} for sensor networks, there is little attention currently to design distributed \ac{ai}-aided \acp{kf}. Design of them for federated learning and edge computing can be a new research direction.
\item {\em Robust training}: Outliers appearing in the SS models, mainly in the observation model  \eqref{eq:ssModelObs}, may substantially affect the training process. The approaches should address this  by providing robust training.
\end{enumerate}



 


\balance
\bibliographystyle{IEEEtran}
\bibliography{IEEEabrv,refs}

\end{document}

%% file: Preamble.tex
\usepackage{bm}
\usepackage{balance}
\usepackage{times}
\usepackage{amsmath,dsfont}
\usepackage{amssymb}
\usepackage{epsfig,verbatim} 
\usepackage{setspace}
\usepackage{color}
\usepackage{cite}
\usepackage{epstopdf}
\usepackage{subcaption}
\usepackage{graphics}
\usepackage{accents}
\usepackage{acronym}
\usepackage[bookmarks,colorlinks]{hyperref}
\usepackage{ulem}
\usepackage{mathtools}
\usepackage{algorithm}
\usepackage{algorithmic}
\usepackage{enumitem}
\usepackage[most]{tcolorbox}
\usepackage{cuted}
\tcbuselibrary{skins}
\usepackage{multicol,lipsum}
\usepackage{wrapfig}
\usepackage{multirow,colortbl,booktabs}
\usepackage{xcolor-material}
\usepackage{tikz}
\usetikzlibrary{fit,positioning,calc,patterns,arrows,shapes.symbols,decorations.shapes,shapes.geometric}
\tikzset{main/.style = {rounded corners,draw,thick,node font={\bf\color{white}},minimum height=10mm,minimum width=40mm}}
\newcommand{\myVec}[1]{{\boldsymbol{#1}}}
\newcommand{\myMat}[1]{{\boldsymbol{#1}}}
\newcommand{\mySet}[1]{\mathbb{#1}}
\newcommand{\myFunc}[1]{\mathcal{#1}}
\newcommand{\Kgain}{\myMat{K}}
\newcommand{\given}[1]{\left|#1\right.}
\newcommand{\dnnParam}{\myVec{\theta}}
\newcommand{\dnnFunc}{\myFunc{F}_{\dnnParam}}
\newcommand{\Ntraining}{N}

\newcommand{\pdf}{p}
\newcommand{\apbm}{\text{\tiny\ac{apbm}}}
\newcommand{\dnn}{\text{\tiny\ac{dnn}}}
\newcommand{\ddm}{\text{\tiny\ac{ddm}}}

\newcommand{\dB}{\left[{\textrm{dB}}\right]}

\definecolor{mypurple}{rgb}{0.910, 0.910, 0.969}
\definecolor{myblue}{rgb}{0.122, 0.435, 0.698}

\definecolor{myGreen}{RGB}{112, 173, 71}
\definecolor{myBlue}{RGB}{36, 124, 185}
 
\acrodef{kf}[KF]{Kalman filter} 
\acrodef{pf}[PF]{particle filter} 
\acrodef{kg}[KG]{Kalman gain}
\acrodef{ekf}[EKF]{extended \ac{kf}}
\acrodef{ukf}[UKF]{unscented \ac{kf}}
\acrodef{ai}[AI]{artificial intelligence} 
\acrodef{dnn}[DNN]{deep neural network} 
\acrodef{cnn}[CNN]{convolutional neural network} 
\acrodef{nn}[NN]{neural network}  
\acrodef{gnn}[GNN]{graph neural network} 
\acrodef{rnn}[RNN]{recurrent neural network} 
\acrodef{fc}[FC]{fully connected} 
\acrodef{mse}[MSE]{mean-squared error}
\acrodef{snr}[SNR]{signal-to-noise ratio}
\acrodef{ss}[SS]{state-space}
\acrodef{ml}[ML]{machine learning}
\acrodef{gru}[GRU]{gated recurrent unit} 
\acrodef{lstm}[LSTM]{long short-term memory} 
\acrodef{ssm}[SSM]{state-space model}
\acrodef{rkn}[RKN]{recurrent Kalman network}
\acrodef{pbm}[PBM]{physics-based model}
\acrodef{apbm}[APBM]{data-augmented physics-based model}
\acrodef{tm}[TM]{true model}
\acrodef{rts}[RTS]{Rauch-Tung-Striebel}
\acrodef{ddm}[DDM]{data-driven model}
\acrodef{vae}[VAE]{variational autoencoder}
\acrodef{armax}[ARMAX]{autoregressive moving-average model}
\acrodef{pinn}[PINN]{physics-informed neural network}
\acrodef{narma}[NARMA]{nonlinear autoregressive moving-average}